\documentclass[journal]{IEEEtran}
\usepackage[linesnumbered,ruled,vlined]{algorithm2e}
\usepackage{booktabs}
\usepackage{threeparttable}
\usepackage{amsmath,amsfonts}

\usepackage{graphicx}
\usepackage{dcolumn}
\usepackage{bm}

\usepackage[utf8]{inputenc}
\usepackage[T1]{fontenc}
\usepackage{mathptmx}
\usepackage{etoolbox}
\usepackage[caption=false,font=normalsize,labelfont=sf,textfont=sf]{subfig}
\usepackage{threeparttable}
\usepackage[colorlinks,
linkcolor=blue,
anchorcolor=blue,
urlcolor=black,
citecolor=blue
]{hyperref}
\usepackage{textcase}
\usepackage{booktabs}

\usepackage{titlesec}
\titlespacing*{\section}
{0pt}{*4}{*1} 
\titlespacing*{\subsection}
{0pt}{*2}{*1} 
\titlespacing*{\subsubsection}
{0pt}{*1}{*1} 

\graphicspath{{figures/}}

\newenvironment{proposition}
{\vspace{0.3em}\noindent\textbf{\textit{Proposition: }}}
{\vspace{0.3em}}

\usepackage{url}

\author{Qihang Chen
\thanks{Qihang Chen is with the School of Information Science and Technology, Fudan University, Shanghai 200433,
China.}
}

\title{Unified Crew Planning and Replanning Optimization in Multi-Line Metro Systems Considering Workforce Heterogeneity}
\begin{document}
\maketitle

\begin{abstract}
Metro crew planning is a key component of smart city development 
as it directly impacts the operational efficiency and
service reliability of public transportation.
With the rapid expansion of metro networks, effective multi-line scheduling and emergency management have become essential 
for large-scale seamless operations.
However, current research focuses primarily on individual metro lines, 
with insufficient attention on cross-line coordination and rapid replanning during disruptions.
Here, a unified optimization framework is presented for multi-line metro crew planning and replanning with heterogeneous workforce.
Specifically, a hierarchical time-space network model is proposed to represent the unified crew action space, 
and computationally efficient constraints and formulations are derived for the crew's heterogeneous qualifications and preferences.
Solution algorithms based on column generation and 
shortest path adjustment are further developed, utilizing the proposed network model.
Experiments with real data from Shanghai and Beijing Metro
demonstrate that the proposed methods outperform benchmark heuristics 
in both cost reduction and task completion, 
and achieve notable efficiency gains by incorporating cross-line operations, 
particularly for urgent tasks during disruptions.
This work highlights the role of global optimization and cross-line coordination 
in multi-line metro system operations, 
providing insights into the efficient and reliable functioning of public transportation in smart cities.

\end{abstract}


\section{Introduction}
\label{section_introduction}
Metro systems are vital to urban transportation, offering high efficiency and large capacity to meet growing mobility demands.
Within the context of metro operations, labor costs account for a significant share of expenses \cite{lin2022metro}. 
Consequently, metro crew planning plays a crucial factor in achieving smooth, cost-effective operations. 
As metro systems continue to expand rapidly, the need for optimized crew planning approaches has become increasingly critical to realize efficient and intelligent metro operations that support the broader goals of smart city development \cite{silva2018towards}.

Existing research on metro crew planning primarily focuses on single-line 
operations \cite{zhou2021integrated,pan2021column,zhou2022metro,jin2022column,feng2023ADMM,xue2023metro}. 
However, current metro systems have evolved into complex multi-line networks with efficient transfer routes. 
Unlike railway and airline systems, where switching routes can take many hours or even days, 
metro crew members can transition between lines within a relatively short time frame. 
This flexibility offers significant opportunities for implementing cross-line operations \cite{qing2019research,sun2022fairness}. 
Such operations enable global optimization by considering the diverse demands and workforce resources of different lines\cite{yong2020study}, 
thereby providing strong practical value. In single-line scenarios, 
researchers often simplify crew characteristics by assuming homogeneity \cite{zhou2010study, zhou2021integrated,feng2023ADMM, zhou2022metro}. 
However, in multi-line settings, crew heterogeneity becomes a critical factor due to the varying technical requirements across line, 
such as differences in train models and signaling systems that demand specific operational qualifications \cite{xie2022hierarchical}. 
Beyond the heterogeneous qualification demands, 
the expanded planning space in multi-line systems necessitates incorporating crew preferences to achieve more considerate crew allocation \cite{jutte2017optimizing}.

Metro systems are highly susceptible to disruptions resulting from factors such as technical malfunctions \cite{deng2020analysis} or unforeseen passenger surges \cite{huan2019early}, making effective disruption management a critical aspect of their operation \cite{wang2024urban}.
Existing research on metro disruption management has primarily focused on rescheduling strategies of train timetables \cite{hou2019energy,zhu2023collaborative} and rolling stock circulation \cite{su2024integrated}, while comparatively little attention has been given to the specific characteristics of metro crew replanning problems.
Specifically, the high-frequency nature of metro operations results in shorter disruption management horizons, demanding more precise time control and greater flexibility in real-time model adjustments. 
Current studies in railway systems often treat crew planning and replanning as separate problems, relying on distinct models and assumptions \cite{cacchiani2014overview}. 
However, this disjointed approach undermines model compatibility, thereby increasing the potential transition costs of switching between planning and replanning models, particularly in scenarios requiring rapid adjustments. 
Therefore, a unified approach that seamlessly integrates crew planning and replanning is of significant importance for enhancing the efficiency of metro disruption recovery. 
Moreover, considering multi-line crew replanning can further improve the effectiveness of disruption recovery by enabling more efficient crew allocation. For instance, when a disruption occurs on one metro line, the multi-line optimization approach could rapidly redeploy available crew members from other lines to minimize the overall adverse impacts.

In summary, existing literature on metro crew planning lacks attention to the special characteristics of multi-line scenarios and the adjustability of replanning strategies.
To address these two gaps, this work proposes a unified crew planning and replanning optimization framework in multi-line metro systems, with particular attention on  workforce heterogeneity.
A unified optimization framework is presented which employs a tailored hierarchical time-space network (HTSN) to model the multi-line action space of both crew planning and replanning, 
formulates both problems as integer programming with derived constraints, and presents respective algorithms for problem solving.
Specifically, the main contributions of this paper are summarized as follows:
\begin{itemize}
\item A novel time-space network model, HTSN, is proposed to represent the entire multi-line planning space, incorporating cross-line operations at both interday and intraday scales. 
The tailored hierarchical structure allows for restricted subnetwork construction, which flexibly converts the model into the replanning space, enabling unified crew management for both normal operations and disruption scenarios.
\item Computationally efficient constraints are derived to ensure the valid pairing of diverse qualifications and duty lists, addressing the unique challenges posed by multi-line scenarios with heterogeneous crew. 
Based on these constraints, integer programming formulations are presented for the crew planning and replanning problems, taking into consideration their distinct operational constraints.
\item Based on the HTSN model, a two-stage column generation (TSCG) method and a fast path adjustment heursitic (FPAH) method are designed to solve the crew planning and replanning problems, 
with a focus on solution quality and computational speed, respectively.
Numerical experiments are conducted using real data from Shanghai Metro and Beijing Metro.
Compared with heuristic benchmarks, the proposed TSCG and FPAH demonstate superior performance in
terms of both operational cost control and train task completion.
The efficiency gains from introducing cross-line operations are also validated, particularly in the completion rate of urgent tasks during disruptions.
\end{itemize}

The remainder of this study is organized as follows. 
Section \ref{section_review} briefly reviews the related literature and compares this work with relevant studies.
Section \ref{section_formulation} provides detailed problem statements and relevant settings.
In Section \ref{section_methodology}, the hierarchical time-space network model, integer programming formulations and the respective solutions are illustrated in detail.
In Section \ref{section_study}, the computational results of numerical experiments and corresponding analysis are presented.
Finally, the conclusion of this study is given in Section \ref{section_conclusion}.

\begin{table*}
\centering
     \begin{threeparttable}    
          \caption{Systematic comparison of related studies.}
          \label{review_table}
          \begin{tabular}{ccccccc}
               \toprule
               Reference               & Problem type & Transportation mode & System scope & Workforce type & Mathematical model & Solution method\\
               \midrule
               Huisman(2007)\cite{huisman2007column}   & CRP  & Railway    & M          & Homo           & SCP   & CG                \\
               Janacek et al.(2017)\cite{janacek2017optimization} & CS      & Railway             & S          & Homo           & NFP    & CG               \\
               Zhou et al.(2021)\cite{zhou2021multi}    & CR      & Airline             & M          & Hetero         & Other & MH \\ 
               Zhou et al.(2021)\cite{zhou2021integrated}    & CP   & Metro                 & S          & Homo           & NFP   & LR              \\
               Breugem et al.(2022)\cite{breugem2022column}     & CRP      & Rail                 & M          & Homo           & IP   & CG               \\
               Zhou et al.(2022)\cite{zhou2022metro}    & CP   & Metro                 & S          & Homo           & MIP   & CG, H              \\
               Pang et al.(2023)\cite{pang2023optimize}    & CS      & Railway             & M          & Hetero           & IP   &  MH  \\
               Feng et al.(2023)\cite{feng2023ADMM}    & CP   & Metro                 & S          & Homo           & NFP   & ADMM, LR         \\
               This study              & CP, CRP   & Metro                 & M          & Hetero         & IP   & CG, H              \\
               \bottomrule
               \end{tabular}
               \begin{tablenotes}
               \scriptsize
               \item \textit{Notes}: CS: crew scheduling; CR: crew rostering; CP: crew planning; CRP: crew replanning;
                S: single line; M: multiple lines; 
               Homo: homogeneous; Hetero: heterogeneous;
               SCP: set covering problem; MIP: mixed integer programming; IP: integer programming; NFP: network flow problem; Other: non-classic mathematical model;
               H: heuristics; CG: column generation; LR: Lagrangian relaxation; MH: meta-heuristics;  ADMM: alternating direction method of multipliers.
               \end{tablenotes}
               \end{threeparttable}
               \end{table*}

\section{Literature Review}
\label{section_review}

\subsection{Crew Scheduling, Rostering and Integrated Planning}
\label{section_review_1}
The crew planning problem aims at allocating specified train service tasks to crew members while satisfying several operational rules, which involves two consecutive stages: crew scheduling and crew rostering. 
The crew scheduling problem involves identifying the least costly set of duties that cover train tasks within the timetable while adhering to specified rules\cite{freling2004decision}. 
Many mathematical models have been proposed for crew scheduling, such as the set covering problem \cite{abbink2005reinventing, han2014constraint} and network flow problem
\cite{janacek2017optimization}.
The time-space network model has been exploited and tailored for crew scheduling 
in many studies\cite{fuentes2019hybrid, xue2023metro,zhou2021integrated,xue2023metro} owing to its flexibility and efficiency in modeling.
Various algorithms have been proposed for scheduling optimization, such as heuristics\cite{ernst2001integrated, zhou2010study}, metaheuristics\cite{fuentes2019hybrid, zhao2022solution} 
and large-scale programming techniques including column generation\cite{abbink2005reinventing, janacek2017optimization,pan2021column,jin2022column,zhou2022metro} and branch-and-price-and-cut\cite{lin2019integrated}.
The crew rostering problem requires determining the most economical set of rosters that encompass given duties and satisfy certain regulations\cite{freling2004decision}.
In this subproblem, several factors involving the rostering quality are often taken into account, 
for instance, robustness of generated rosters \cite{lusby2012column}, 
fairness of working conditions \cite{nishi2014two} 
and balancing of workload \cite{zhou2021multi}.
To reach global optimality, a few studies have also developed integrated planning approaches to optimize crew scheduling and rostering simultaneously\cite{ernst2001integrated,souai2009genetic,zhou2021integrated,feng2023ADMM}.
Time-space network has also been considered for modeling the integrated problem.
Zhou et al.\cite{zhou2021integrated} used a multilayer time-space network formulation along with duty time window design to model metro crew planning with flexible duty interavls, 
and Feng et al.\cite{feng2023ADMM} proposed a three-dimensional time-space-state network that dynamically updates behavior states for metro crew members in the generated rosters.

\subsection{Cross-Line Operations and Workforce Heterogeneity}
\label{section_review_2}
In multi-line transportation systems, many cross-line coordination strategies have been studied, 
such as the design of cross-line express trains for timetabling \cite{zhang2024integrated} and service replanning \cite{yang2020service}, the optimization of cross-line train routing schemes for travel fairess enhancement \cite{sun2022fairness}, etc.
In the context of crew planning, Pang et al.\cite{pang2023optimize} incorporated cross-line deadhead choices to railway crew scheduling to alleviate workforce shortage.
Zhao et al.\cite{zhao2022solution} adopted a cooperation mode between different crew depots that allows for cross-line duty scheduling for high-speed railway crew units.
With the increasing complexity of transportation systems, issues associated with heterogeneous workforce management have also emerged\cite{xie2022hierarchical}.
To address the diverse individual preferences, 
Quesnel et al.\cite{quesnel2020improving} focused on preferred legs and  offperiods to enhance airline rostering solution quality, and Zhou et al.\cite{zhou2021multi} included the preferred flights and vacations in the optimization objectivefor improving the crew's satisfaction rate.
In terms of heterogeneous working qualifications, Pang et al.\cite{pang2023optimize} constructed different crew pools for each occupation's available crew members, and
Wen et al.\cite{wen2022individual} modeled cabin crews individually instead of team modelling to handle heterogenous manpower.

\subsection{Disruption Management and Crew Replanning}
\label{section_review_3}
The real-time operations of metro and rail systems are unavoidably subject to unexpected disruptions \cite{cacchiani2014overview}.
Disruption management can be examined from pre-disruptive and post-disruptive perspectives\cite{wang2024urban} and involves key approaches like timetable adjustment, rolling stock rescheduling and crew replanning \cite{zhang2021metro}. 
The crew replanning problem aimes at finding a new feasible crew roster given the updated timetable, updated rolling stock circulation and the current crew roster \cite{cacchiani2014overview}. 
Many studies use set covering models to cover as many updated tasks as possible \cite{huisman2007column,rezanova2010train,potthoff2010railway,sato2011real}. 
Rezanova and Ryan \cite{rezanova2010train} and Potthoff et al. \cite{potthoff2010railway} include the option of canceling
tasks with large canceling penalties for feasible rescheduling results. To enable real-time replanning, Sato and Fukumura\cite{sato2011real} consider only the duties
affected by the disruption to save computational time. 
It is noted that current research on crew replanning is predominantly focused on railways and airlines, with limited studies addressing metro systems.

\subsection{Comparison with Related Studies}
\label{section_review_4}

For a systematic comparison, the detailed characteristics of related studies are summarized in Table \ref{review_table}. 
Notably, this study is the first attempt to investigate operations in multi-line metro systems, 
and to address crew planning and replanning within a unified framework. 
The proposed approach incorporates heterogeneous preferences and qualifications, 
aligning partially with Zhou et al.\cite{zhou2021multi} and Pan et al.\cite{pan2021column}, while adapting their formulation techniques to metro-specific scenarios.
The replanning model shares similar objectives with related studies\cite{huisman2007column, breugem2022column}, yet introduces significant adjustments to suit metro disruption management. Specifically, this work employs shorter replanning horizons and imposes stricter continuity constraints to accommodate the high-frequency nature of metro operations.

The proposed hierarchical time-space network draws on common design principles observed in \cite{zhou2021integrated, zhou2022metro, feng2023ADMM},  but emphasizes different aspects. Feng et al. \cite{feng2023ADMM} underlined the additional state dimension to model exact state transitions of crew members, while Zhou et al. \cite{zhou2022metro} focused on the multilayer structure to distinguish various duty types and rostering schemes. 
The proposed network design most closely aligns with \cite{zhou2021integrated} in terms of state and arc representations. 
However, to address the challenges posed by multiple metro lines and disruption recovery, 
this work extends the model in \cite{zhou2021integrated} by introducing additional vertex and arc types tailored to these scenarios, as detailed in Section S.I in the supplementary material. 
Furthermore, this work organizes the network into a novel hierarchical structure to facilitate efficient restricted subnetwork construction during disruptions.

\section{Problem Formulation}
\label{section_formulation}

The basic elements in the studied crew planning and replanning problems are first described.
Next, the formulation of workforce heterogeneity and cross-line operations are illustrated.
The respective optimization objectives and operational rules are further described.

\subsection{Basic Settings}

The basic elements in the studied crew planning and replanning problems are described as follows:

\textbf{(i)} Planning Horizon and Granularity:
Different from airlines or railways, the planning horizon of metro systems is typically set on a daily basis,
and a high level of temporal planning granularity is required due to the frequent short-distance travel demands.
In this study, the planning horizon is set to several consecutive working days, denoted as $D = \{d_1, d_2, ..., d_{n_d} \}$.
Following previous works\cite{xue2023metro,feng2023ADMM,zhou2021integrated,zhou2022metro}, this work discretizes the daily time horizon into basic time units, 
with each time unit represent one minute.
Specifically, let $[T_b, T_e]$ denote the discretized time horizon of $T_e-T_b+1$ minutes within each day.

\textbf{(ii)} Line, Depot and Transfer Station:
This work considers a metro system with multiple double-track lines, denoted as $L = \{l_1, l_2, ..., l_{n_l}\}$.
Within each metro line,
depots are the stations that serve as the origin and destination for trains tasks, as well as locations where crew members sign in, sign out, take breaks and have meals.  
Two terminal stations at both ends of a metro line are set as depots, denoted as $O_l= \{o_{la}, o_{lb}\}$ for line $l \in L$.
For two intersecting lines $l_1$ and $l_2$, denote the transfer station as $f_{l_1, l_2}$.

\textbf{(iii)} Train Service Task:
A train service task refers to the process in which a crew member, serving as a qualified operator, facilitates the operation of a train traveling from one terminal station to another.
Train service tasks are generated according to the train timetables, and could be cancelled due to insufficient qualified workforce or real-time disruptions.
Specifically, let $K$ denote the set of all train service tasks.
For each task $k \in K$,
a quintuple $(o_k, t_k, o_k', t_k', \lambda_k)$ is specified 
representing the departure station, departure time, arrival station, arrival time and cancelling penalty of task $k$, respectively.

\label{section_formulation_duty}
\textbf{(iv)} Duty and Duty List:
A duty refers to the daily working arrangement assigned to a crew member, encompassing activities such as signing in, completing tasks and taking breaks within a specific time frame.
To determine the duty time frames, this work adopts the sliding time window design proposed by Zhou et al. \cite{zhou2021integrated}, 
which partitions the daily time horizon into potential duty frames using a fixed-length window $H$ and a sliding interval $h$.
Specifically, let $U = \{u_1, u_2, ..., u_{n_u}\}$ denote the set of all possible duty frames within each day, where each $u$ spans $[a_u, a_u + H]$. 
Each crew member is assigned to a single duty frame per day, and the sequence of these daily duties over the planning horizon forms one duty list, representing the planning outcome for the individual. 
The duty lists for all crew members collectively constitute the roster, which serves as the final output of the crew planning process.

\begin{figure}[!t]
     \centering
     \includegraphics[height=0.24\textheight]{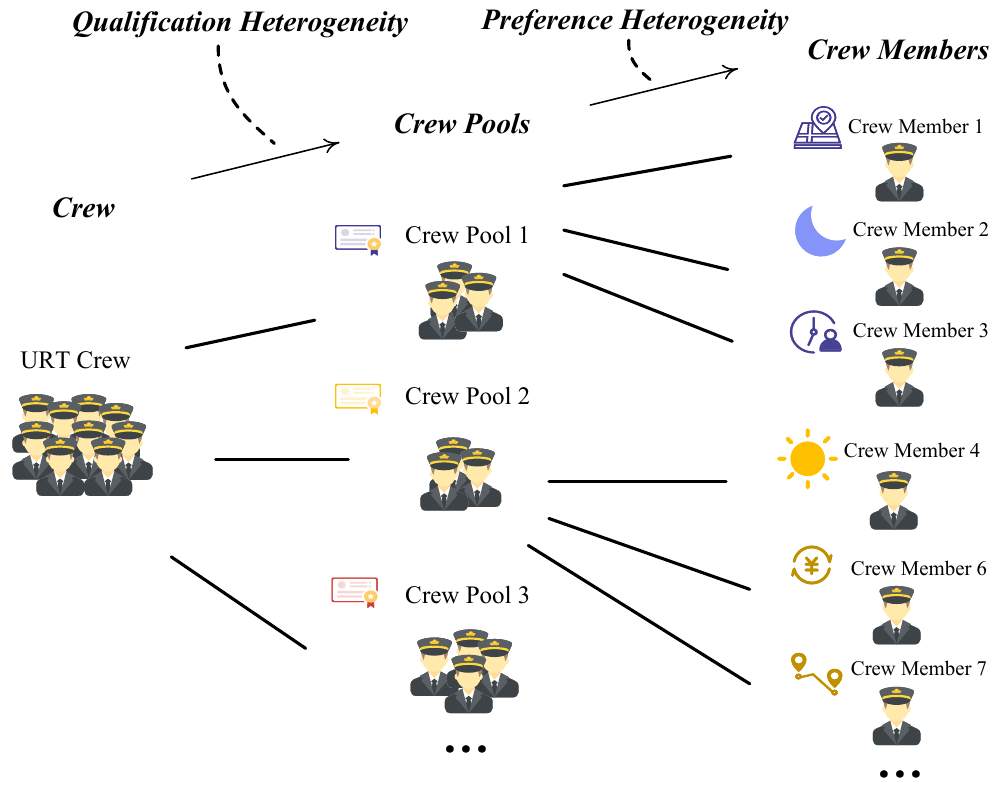}
     \caption{
          Formulation of qualification heterogeneity and preference heterogeneity among metro crew members.
          }
     \label{workhetero}
\end{figure}

\subsection{Formulation of Workforce Heterogeneity}
\label{section_formulation_heterogeneity}
This study considers a heterogeneous crew of  $n_r$ members, denoted as $R = \{r_1, r_2, ..., r_{n_r}\}$, each characterized by distinct attributes.
To capture workforce heterogeneity, this work approachs it from two key dimensions: qualification heterogeneity and preference heterogeneity. 
Qualification heterogeneity refers to the variation in the operating skills of crew members, which determines their ability to perform tasks on specific metro lines. 
Preference heterogeneity, on the other hand, encompasses the individual inclinations of crew members, such as preferences for tasks on particular metro lines, and choices regarding sign-in and sign-out depots. 
This work employs distinct strategies to mathematically model these two aspects of workforce heterogeneity.

Due to explicit requirements, qualification heterogeneity is modeled as hard constraints in the optimization framework. Following the approach of Pang et al. \cite{pang2023optimize}, 
this work first defines the qualification set $Q = \{q_1, q_2, ..., q_{n_q}\}$, representing all possible combinations of metro lines.
Mathematically, we have 

$Q = \{\tilde{L} | \tilde{L} \subseteq L, \tilde{L} \neq \phi\}$.
The whole crew is then divided into $n_q$ crew pools, denoted as $P = \{P_{q_1}, P_{q_2}, ..., P_{q_{n_q}}\}$, as illustrated in Fig. \ref{workhetero}.
Duty lists requiring specific line assignments should only be matched to crew members with the necessary qualifications.

In contrast to qualification heterogeneity, crew preferences are more flexible and are therefore incorporated into the optimization objective. To account for this, 
preference violation penalties are introduced by quantifying the extent to which a duty list deviates from a crew member's preferences. This study focuses on crew preferences related to sign-in and sign-out depots, a common preference type in the literature \cite{quesnel2020improving, zhou2021multi}. 
Specifically, each crew member $r$ has a set of prefered depots, denoted as $O_r$.
Additionally, $\lambda_o$ is defined
as the penalty cost incurred when a crew member is assigned to sign in or out at an unpreferred depot, providing a straightforward method for quantifying preference violations.

\begin{figure}[!t]
     \centering
     \includegraphics[height=0.24\textheight]{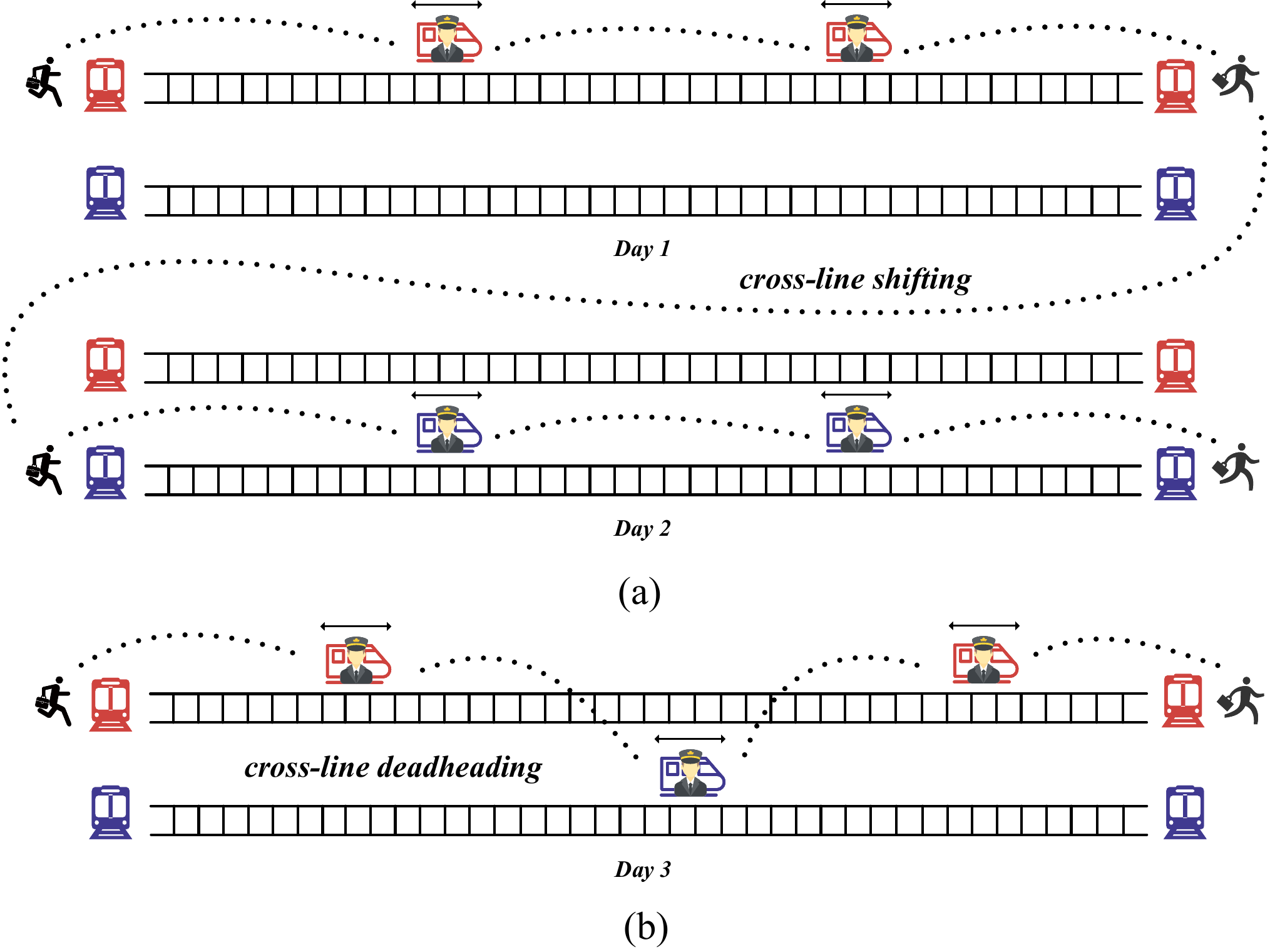}
     \caption{
          Illustration of metro crew cross-line operations: (a) interday cross-line shifting and (b) intraday cross-line deadheading.
     }
     \label{crossline}
\end{figure}

\subsection{Cross-Line Operations}
\label{section_formulation_crossline}

This work introduces cross-line operations for crew members qualified on multiple lines, focusing on both interday and intraday scales.
Interday cross-line operations allow crew members to switch working lines between different days within a planning horizon. For example, a crew member qualified on lines $l_1$ and $l_3$ might work on line $l_1$ on the first day and switch to line $l_3$ the following day, as shown in Fig. \ref{crossline} (a).
Intraday cross-line operations enable crew members to deadhead to other lines to address urgent needs within a single working day. For instance, if line $l_3$ experiences increased demand during the day, a crew member qualified on both $l_1$ and $l_3$ could deadhead to line $l_3$, work there for a few hours, and then return to line $l_1$ once the demand is addressed, as illustrated in Fig. \ref{crossline} (b). 
Since cross-line express trains \cite{yang2020service} are not considered in this study, it is assumed that intraday cross-line deadheading is implemented via existing metro train services. 
Let $t^d_{k_1, k_2}$ denote the time cost of a deadhead task via task $k_1$ and $k_2$, then we have 
$t^d_{k_1, k_2} = t_{k_2}' - t_{k_1}.$

\begin{figure*}
     \centering
     \includegraphics[height=0.24\textheight]{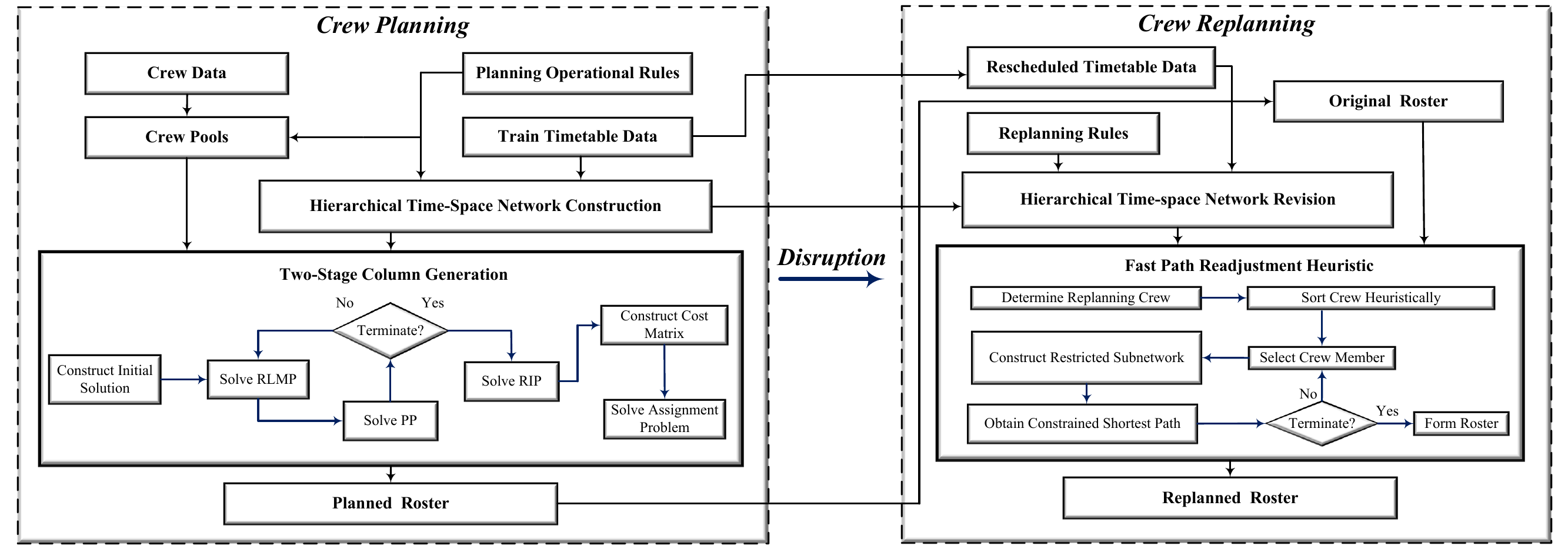}
     \caption{
          Complete pipeline of the unified crew planning and replanning optimization framework.
     }
     \label{pipeline}
\end{figure*}

\subsection{Objective and Operational Rules}
\label{section_formulation_rules}

For the crew planning problem, this work quantifies crew labor input, task cancellation penalties, and preference violation penalties as distinct cost terms, and set the overall cost minimization as the optimization objective. 
To represent labor input, let $c_w$ and $c_r$
be the costs per time unit for performing train service tasks and taking non-working actions, respectively.
To ensure the model's practicality, this work adopts the following operational rules for feasible crew planning.

\begin{itemize}
     \item Day-off rule: Crew members must take at least $n_{df}$ days off for rest during a single planning horizon.
     \item Sign-in and sign-out rule: Crew members must begin each working day with a sign-in task and conclude with a sign-out task, taking $t_{si}$ and $t_{so}$ time units, respectively.
     \item Train service rule: Each train service task should be performed by at most one crew member.
     \item Working time rule: The daily working time, refering to the period from the first sign-in time unit to the last sign-out time unit, must fall within $[t_{min}, t_{max}]$.
     \item Rest break rule: Crew members must take at least $t_{rt}$ time units off for rest between performing two consecutive train service tasks.
     \item Meal break rule: Crew members must be allotted $t_{ml}$ time units for a meal break within their duty $u$, during the time period $[a_u + t_{mb}, a_u + t_{me}]$, on each working day.
     \item Qualification rule: Crew members must perform train service tasks on metro lines that match their qualifications.
     \item Deadhead rule: Crew members must be assigned no more than $n_{tf}$ deadhead tasks within one planning horizon.
\end{itemize}

Compared with crew planning, the crew replanning problem focuses on addressing new task demands while avoiding large-scale changes to the original roster\cite{cacchiani2014overview}, thus introducing additional constraints.
To account for the fact that many crew members are already on duty when real-time replanning is implemented, and
they expect to work with continuity in the same period as from the initial planning\cite{malucelli2019delay,breugem2022column}, 
the following rules should be additionally satisfied for the crew replanning problem.

\begin{itemize}
     \item Replanned duty rule: Crew members should be assigned duties with time frames consistent with their original ones during replanning.
     \item Replanned task rule: Crew members should be assigned tasks that align seamlessly with the original ongoing tasks in terms of both time and location during replanning.
\end{itemize}

All the notations mentioned for problem formulation are summarized in Table S.I in the supplementary material.

\section{Methodology}
\label{section_methodology}

The hierarchical time-space network (HTSN) model, the integer formulations for crew planning and replanning problems, 
as well as the respective solution algorithms are described.
Fig. \ref{pipeline} illustrates the entire pipeline of the unified optimization framework.

\subsection{Hierarchical Time-Space Network}

\begin{figure*}[!t]
     \centering
     \includegraphics[height=0.36\textheight]{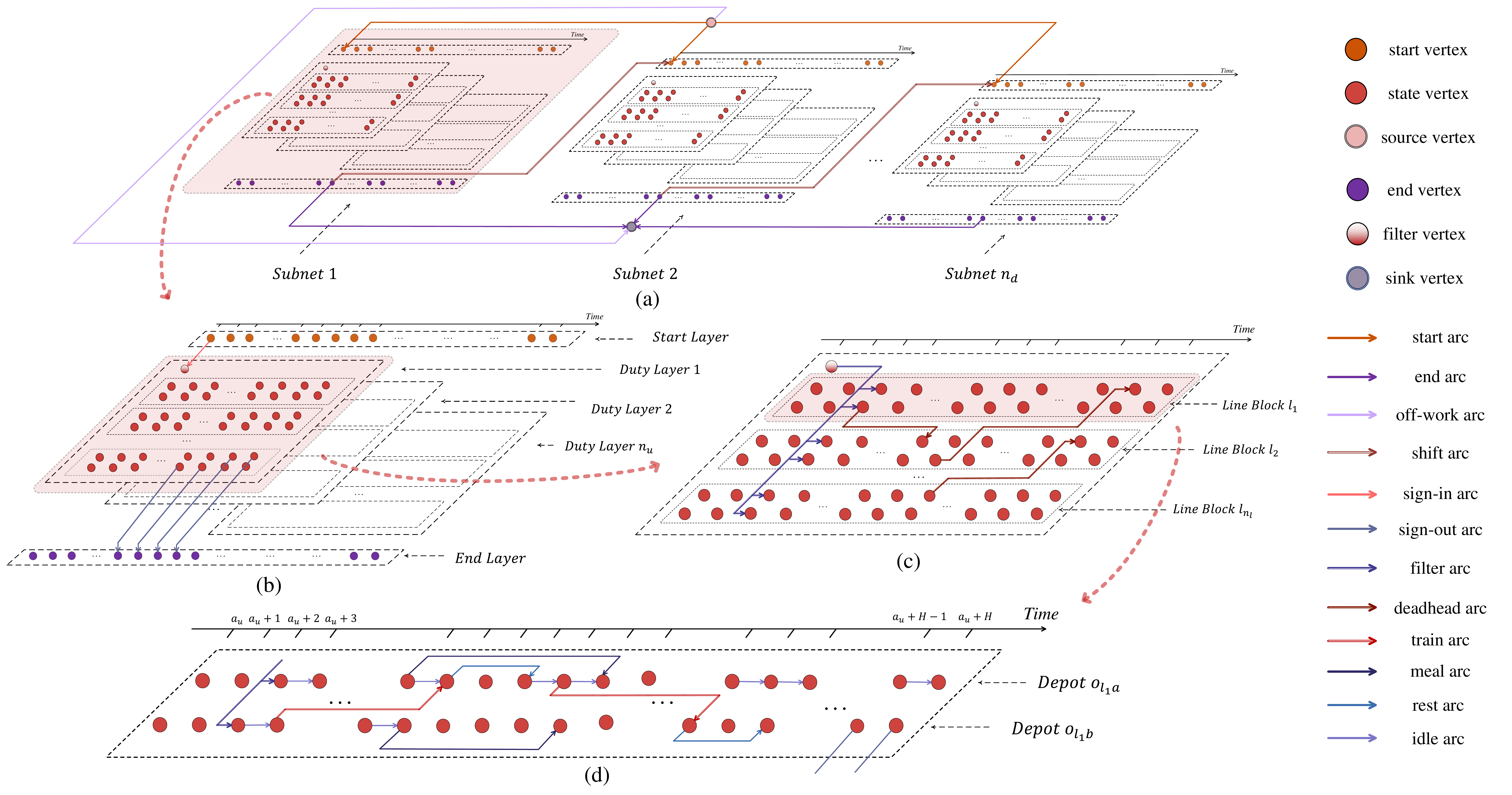}
     \caption{
          Illustration of the hierarchical time-space network (HTSN) model: 
          (a) overall network structure, 
          (b) daily subnetworks, 
          (c) duty layers, and 
          (d) line blocks.
     }
     \label{SMTSN}
\end{figure*}

This work proposes a hierarchical time-space network (HTSN) model, 
which is designed to represent the entire potential action space for crew duty rosters while satisfying several operational rules.
The proposed HTSN model draws on common design principles with related studies \cite{zhou2021integrated, zhou2022metro, feng2023ADMM} and emphasizes distinct aspects.
Specifically, vertices in HTSN represent crew members' different states during planning, 
and directed arcs represent several kinds of actions that crew members can take, such as signing in, performing train service tasks, taking meals, etc. 
To enhance structural adaptability, HTSN is designed as a four-tier hierarchical structure, comprising the line block, duty layer, daily subnetwork, and hierarchical network from the bottom to the top, as depicted in Fig. \ref{SMTSN}.

A line block represents the search space for the crew to take actions on a specific metro line within a duty frame on a given day, as illustrated in Fig. \ref{SMTSN} (d).
Let $V^{d, u, l}_{block}$ denote the line blcok for line $l$ within duty frame $u$ on day $d$, where each state vertice $v^{d, u, l}_{t, o} \in V^{d, u, l}_{block}$ represents that a crew member is located at depot $p$ at time $t$.
Withing each line block, train arcs $v^{d, u, l}_{t_k, o_k} \to v^{d, u, l}_{t_k', o_k'}$ are constructed according to the train timetable, each followed by one a rest arc $v^{d, u, l}_{t_k', o_k'} \to v^{d, u, l}_{t_k' + t_{rt}, o_k'}$ to satisfy the rest break rule 
Meal arcs $v^{d, u, l}_{t, o} \to v^{d, u, l}_{t + t_{ml}, o}$ are added adhering to the meal break rule, and necessary idle arcs $v^{d, u, l}_{t, o} \to v^{d, u, l}_{t + t', o}$ are inserted between certain unconnected state vertices to ensure time-space connectivity.

A duty layer is the multi-line search space for 
for one duty frame on a given day, consisting of $n_l$ line blocks horizontally aligned along the temporal dimension, as illustrated in Fig. \ref{SMTSN} (c). 
Let $V^{d, u}$ denote the duty layer for duty $u$ in day $d$.
Considering the qualification heterogeneity, a filter vertex $v^{d, u}_{ft}$ is added
for each duty layer as the logistic control unit for checking valid qualifications,
and fiter arcs $v^{d, u}_{ft}\ \to v^{d, u, l}_{a_u+t_{si}, o}$ are inserted to connect the filter vertex to state vertices on qualified metro lines,
ensuring that crew members could only sign in at qualified depots.
Meanwhile, intraday cross-line operations are allowed by inserting deadhead arcs $v^{d, u, l_1}_{t_{k_1}, o_{k_1}} \to v^{d, u, l_2}_{t_{k_2}', o_{k_2}'}$
between line blocks, where $k_1$ and $k_2$ are a pair of train services tasks on $l_1$ and $l_2$ with shortest extra waiting time..

A daily subnetwork, as depicted in Fig. \ref{SMTSN} (b), represents the entire daily planning space consists of $n_u$ independent duty layers, one start layer and one end layer,
denoted as $V_{subnet}^d = \bigcup_{u \in U}V_{layer}^{d,u} \cup V_{st}^d \cup V_{ed}^d$.
A start vertex $v^{d}_{st, t} \in V^d_{st}$ indicates that a member starts working at time $t$ in day $d$.
and sign-in arcs $v^{d}_{st, a_u} \to v^{d, u}_{ft}$ are constructed between start vertices and filter vertices. 
Similarly, sign-out arcs $v^{d, u, l}_{t, o} \to v^{d}_{ed, t+t_{so}}$ are inserted between state vertices and end vertices, where $t+t_{so}$ should fall into $[a_u + t_{min}, a_u + t_{max}]$ in accordance with the working time rule.

The entire hierarchical network representation begins with a source vertex and ends with a sink vertex, 
representing the start and end of a planning horizon, respectively, as illustrated in Fig. \ref{SMTSN} (a).
Along with the source and sink vertices, we have the network vertex set 
$V = \bigcup_{d \in D}V^d_{subnet} \cup \{v_{sc}, v_{sk}\}$.
The source and sink vertices are connected with the start and end vertices, respectively, using start arcs $v_{sc} \to v^d_{st, t}$ and end arcs $v^d_{ed, t} \to v_{sk}$.
An off-work arc $v_{sc} \to v_{sk}$ is added for non-working member during planning. 
Each pair of daily subnetworks are connected by shift arcs  $v^{d}_{ed, t} \to v^{d'}_{st, t'}$. 
Flexible shift arcs, combined with filter arcs, facilitate interday cross-line operations by enabling crew members to switch lines across different days.

Traversing an arc in the network means that a certain action has been taken, resulting in a corresponding cost and a spatio-temporal transition for the crew member.
A path from the source vertex to the sink vertex then reveals all the actions that a crew member takes as well as the time-space trajectory during an entire planning horizon.
Consequently, such a network path corresponds to a duty list for one crew member, as described in Section \ref{section_formulation_duty}.

The notations of the vertices and arcs in HTSN are summarized in Tables S.II and S.III in the supplementary material.

\subsection{Integer Programming Formulation}

Utilizing paths in HTSN, this work formulates both the crew planning and replanning as integer programming problems.
To address the unique requirement of heterogeneous qualifications in multi-line scenarios, this work first proposes a type of constraints based on perfect matching 
on bipartite graph to reduce the programming scale and complexity.
The integer programming formulations for crew planning and replanning are further presented.
To satisfy the replanning rules, a restricted subnetwork construction technique is designed, demonstrating the high flexibility and adaptability of the proposed hierarchical network model.

\subsubsection{Perfect Matching for Qualification Heterogeneity:}
Given the hierarchical network, duty list candidates can be generated by searching feasible paths. 
Selecting duty lists from candidates then requires additional consideration of heterogeneous qualification matching. Here the qualification rule is formulated with a bipartite graph structure. 
Specifically, let crew members and duty lists form two vertex groups, and let arcs represent 
compatibility between a crew member's qualifications and a duty list's requirements. 
The satisfaction of qualification rule is then transformed into finding a perfect matching on the bipartite graph. 
To guarantee the existence of a perfect matching,
the \textit{Hall's condition} in the classic Hall's theorem\cite{cameron1994combinatorics} is commonly adopted.
The original Hall's condition, however, imposes $O(2^{n_r})$ constraints which could result in computational infeasibility when the crew size is large.
To address this issue,
this study proposes a mathematically equivalent constraint leveraging the feature of the constructed graph.
Given network $G$, let $\Gamma$ be the set of all viable network paths. 
For each path $\gamma \in \Gamma$, 
let decision variable $x_\gamma \in \{0, 1\}$ represent whether path $\gamma$ is selected, 
and parameters $m_\gamma^q \in \{0, 1\}$ to represent whether the line requirements of path $\gamma$ equal qualification $q$.
Next, let $E (q) = \{q' | q \subseteq q', q' \in Q\}$ be the superset group of qualification $q$, and $\mathbf{E} = \{E(q)|q \in Q\}$. 
Finally, for any subset $\tilde{\mathbf{E}} \subseteq \mathbf{E}$, let $S(\tilde{\mathbf{E}}) = \bigcup_{E \in \tilde{\mathbf{E}}}E$ denote the superset extended group for $\tilde{\mathbf{E}}$, and $\mathbf{S} = \{S(\tilde{\mathbf{E}})|  \tilde{\mathbf{E}} \subseteq \mathbf{E},\tilde{\mathbf{E}} \neq \phi \}$.
In turn, one proposition is presented as follows.

\begin{proposition}
\label{proposition}
There exists a perfect matching between the selected paths and crew members 
if and only if the following constraint for each $S \in \mathbf{S}$ is satisfied: $\sum_{\gamma \in \Gamma}\sum_{q \in S}m_{\gamma}^{q}x_\gamma \le \sum_{q \in S}\left |P_q \right |$.
\end{proposition}
\vspace{5pt}

Based on this proposition, the qualification rule can be satisfied with $O(2^{n_q})$ constraints.
Since in general we have $n_q \ll n_r$, this formulation
is acceptable for model solving in terms of constraint complexity.
A rigorous proof for the proposition is provided in Section S.II in the supplementary material.

\subsubsection{Integer Programming for Crew Planning:}
With the matching-based constraints, the integer programming formulation for crew planning is presented.
For each path $\gamma \in \Gamma$, let $A_\gamma$ denote the set of traversed arcs.
The labor costs then consist of the arc costs $c(u, v)$, where $u \to v \in A_\gamma$.
For the preference violation penalties, 
let decision variables $y_\gamma^r \in \{0, 1\}$ denote whether path $\gamma$ is assigned to crew member $r$,
and let $O_\gamma$ denote the set of depots for each sign-in or sign-out along path $\gamma$.
To formulate constraints, 
let $n_\gamma^{si}$ and  $n_\gamma^{dh}$ denote the number of sign-in arcs and deadhead arcs traversed in this path, 
and $n_\gamma^{ml, d}$ denote the number of meal arcs traversed in the day $d$.
For each train service task $k$, let $\kappa_\gamma^k \in \{0, 1\}$ denote whether it is covered by path $\gamma$.
The integer programming problem is formulated as follows:
\begin{align}
 \mathbf{P}: \mathrm{Minimize} \ &\sum_{\gamma \in \Gamma}\sum_{u \to v \in A_\gamma}c(u, v) x_\gamma   + \sum_{k\in K}\lambda_k& \nonumber \\
 & + \sum_{r \in R}\sum_{\gamma \in \Gamma}\sum_{o \in O_\gamma}y_\gamma^r\delta(o \notin O_r)\lambda_o& \label{object} \\ 
\mathrm{subject \ to} \ &\sum_{\gamma \in \Gamma}\sum_{q \in S}m_{\gamma}^{q}x_\gamma \le \sum_{q \in S}\left |P_q \right |, \ \ \forall S \in \mathbf{S}&\label{c1} 
\end{align}
\begin{align}
& \sum_{\gamma \in \Gamma}\kappa_{\gamma}^k x_\gamma \le 1, \ \ \forall k \in K \label{c2}& \\
& \sum_{r \in R}y_\gamma^r = x_\gamma, \ \ \forall \gamma \in \Gamma& \label{c3} \\
& \sum_{\gamma \in \Gamma}y_\gamma^r = 1, \ \ \forall r \in R &\label{c4} \\
& m_\gamma^q y_\gamma^r\le \delta(q \subseteq q^r), \ \ \forall \gamma \in \Gamma, r \in R, q \in Q &\label{c_r} \\
& n_\gamma^{si} \le n_d - n_{df}, \ \ \forall \gamma \in \Gamma &\label{c5}  \\
& n_\gamma^{dh} \le n_{tf}, \ \ \forall \gamma \in \Gamma& \label{c6}\\
& n_\gamma^{ml, d} = 1, \ \ \forall \gamma \in \Gamma, \forall d \in D& \label{c7} \\
& x_\gamma \in \{0, 1\}, \ \ \forall \gamma \in \Gamma&\label{c8} \\
& y_\gamma^r \in \{0, 1\},\ \ \forall \gamma \in \Gamma, \forall r \in R& \label{c9} 
\end{align}
The objective function (\ref{object}) minimizes the total cost including the labor costs, preference violation penalties and train service canceling penalties, where $\delta(\cdot)$ is the Kronecker function that returns a boolean value.
Constraints (\ref{c1})-(\ref{c2}), (\ref{c5})-(\ref{c7}) satisfy the  qualification rule, train service rule, day-off rule, deadhead rule and meal break rule, respectively.
Constraints (\ref{c3})-(\ref{c_r}) guarentee that each selected path is associated with one qualified crew member and every member is assigned.

\subsubsection{Restricted Subnetwork Construction: }
\begin{figure*}[!t]
     \centering
     \includegraphics[height=0.16\textheight]{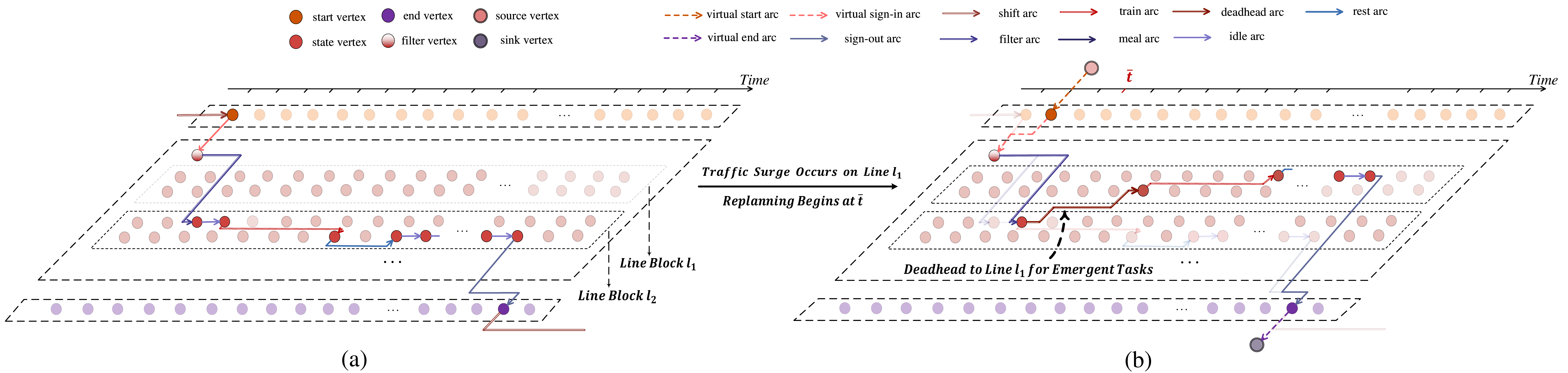}
     \caption{
          Illustration of the restricted subnetwork construction technique, which adjusts the planning space for rapid replanning:
           (a) the original network path in HTSN on day $\bar{d}$, 
           and (b) the replanned network path in the restricted subnetwork on day $\bar{d}$.
     }
     \label{subnet}
\end{figure*}

For the crew replanning problem, the planning search space has to be adjusted to accommodate dynamic changes and meet additional rules described in Section \ref{section_formulation_rules}. 
This work introduces the restricted subnetwork construction technique, illustrated in Fig. \ref{subnet}, which satisfies the replanning duty and task rules solely by modifying the proposed hierarchical time-space network.

Specifically, 
let $\bar{d}$ and $\bar{t}$ to be the replanning day and replanning start time when disruption occurs. Let $\bar{\Gamma}$ be the set of paths that traversed working arcs during day $\bar{d}$ in the original roster, and $r_{\gamma}$ be the assigned crew member for each $\gamma \in \bar{\Gamma}$.
Thanks to the high modeling accuracy of the hierarchical network, it is straightforward to obtain the selected duty $u_\gamma$ in day $\bar{d}$ and the traversing arc $v_{\gamma}' \to v_{\gamma}$ at time $\bar{t}$ given path $\gamma  \in \bar{\Gamma}$.
To ensure a valid connection with the previously established trajectory, a virtual sign-in arc $v^{\bar{d}}_{st, t_{v_{\gamma}} - t_{si}} \to v_{ft}^{\bar{d}, u_\gamma}$ is constructed 
and the corresponding filter arcs $v_{ft}^{\bar{d}, u_\gamma} \to v^{\bar{d}, u_\gamma, l}_{{t_{v_{\gamma}}}, o_{\gamma}}$ are inserted into duty layer $V^{\bar{d}, u_\gamma}_{layer}$ to readjust the time-space network.
The restricted subnetwork tailored for path $\gamma$ as $G^\gamma = (V^\gamma, A^\gamma)$ is then constructed, where $V^\gamma = V^{\bar{d}, u_\gamma}_{layer} \cup V_{ed}^{\bar{d}} \cup \{v_{sc}, v_{sk}, v^{\bar{d}}_{st, t_{v_{\gamma}} - t_{si}}\}$, $A^\gamma = \{u \to v |  u, v \in V^\gamma, u \to v \in A\}$.
After constructing a restricted subnetwork for each crew member, we can now search for valid network paths from source to sink to generate the replanned roster while maintaining spatiotemporal continuity with the original plan.

\subsubsection{Integer Programming for Crew Replanning: }
After constructing restricted subnetwork $G^\gamma$ for each $\gamma \in \bar{\Gamma}$, let $\Omega^\gamma$ denote the set of all viable network paths in $G^\gamma$.
Let $\bar{K}$ denote the replanned task set given the new timetable, and let $\vartheta_k$ and $\bar{\lambda}_k$ be the required number of crew members for conducting task $k$ and the corresponding new canceling penalty, respectively.
The replanning problem is then formulated as follows:
\begin{align}
 \mathbf{RP}: \mathrm{Minimize} \ &\sum_{\gamma \in \bar{\Gamma}}\sum_{\omega  \in \Omega^\gamma }\sum_{u \to v \in A^\omega}c(u, v) x_\omega   + \sum_{k\in \bar{K}}\vartheta_k\bar{\lambda}_k& \nonumber \\
 & + \sum_{\gamma \in \bar{\Gamma}}\sum_{\omega \in \Omega^\gamma}\sum_{o \in O_\omega}x_\omega\delta(o \notin O_{r_\gamma})\lambda_o& \label{robject} 
\end{align}
\begin{align}
 \mathrm{subject \ to} \ &\sum_{\gamma \in \bar{\Gamma}}\sum_{\omega \in \Omega^\gamma}\sum_{q \in S}m_{\omega}^{q}x_\omega \le \sum_{q \in S}\left |P_q \right |, \ \ \forall S \in \mathbf{S}&\label{rc1}\\
& \sum_{\omega \in \underset{\gamma \in \bar{\Gamma}}{\bigcup}\Omega^\gamma}\kappa_{\omega}^k x_\omega \le \vartheta_k, \ \ \forall k \in \bar{K} \label{rc2}& \\
& m_\omega^q x_\omega\le \delta(q \subseteq q^r_\gamma), \ \ \forall \omega \in \bigcup_{\gamma \in \bar{\Gamma}} \Omega^\gamma , q \in Q &\label{rc4} \\
& n_\gamma^{ml, \bar{d}, \bar{t}} + n_\omega^{ml} = 1, \ \ \forall \omega \in \Omega^\gamma, \gamma \in \bar{\Gamma}& \label{rc5} \\
& \sum_{\omega \in \Omega^\gamma}x_\omega = 1, \ \ \forall \gamma \in \bar{\Gamma}& \label{rc3}  \\
& x_\omega \in \{0, 1\}, \ \ \forall \omega \in \bigcup_{\gamma \in \bar{\Gamma}}\Omega^\gamma&\label{rc6}
\end{align}
Compared with the original function (\ref{object}), the replanning objective (\ref{robject}) changes the term of service canceling penalties to incorporate the specific demands of replanned tasks.
Constraints (\ref{rc2}) relaxes the original train service rule by allowing multiple crew members to operate pressing tasks during disruption. 
The original constraints (\ref{c5})-(\ref{c6}) for day-off and deadhead rules are removed, but the meal break rule still preserves through constraints (\ref{rc5}), where $n_\gamma^{ml, \bar{d}, \bar{t}}$ denotes the number of meals already taken in the original plan before $\bar{t}$ in day $\bar{t}$ .
Unique path selection is imposed for the heterogeneous crew member with constraints (\ref{rc3}).

This work proposes a two-stage column generation (TSCG) method to solve the crew planning problem $\mathbf{P}$.
For solving problem $\mathbf{RP}$, this work designs a fast path adjustment heuristic (FPAH) on the basis of HTSN restricted subnetwork construction.
The details of the solution methods are detailed in the Methods section.

\subsection{Two-Stage Column Generation}
\label{section_algorithm_tso}

This work proposes a two-stage column generation method to solve the crew planning problem $\mathbf{P}$.
The complexity of solving problem $\mathbf{P}$
primarily arises from the enoumous amount of potential path candidates in $\Gamma$ and the coupling relationship between decision variables $x_\gamma$ and $y_\gamma^r$. 
To address the first challenge, column generation is leveraged, a commonly adopted technique for problems where enumerating all variables is computationally infeasible\cite{barnhart1998branch,desaulniers2006column}.
For the second challenge, a model decomposition strategy is applied, decoupling problem $\mathbf{P}$ into a network flow problem $\mathbf{P_1}$ and an assignment problem $\mathbf{P_2}$.
$\mathbf{P_1}$ and $\mathbf{P_2}$ are solved sequentially in a two-stage approach, utilizing the techinique of column generation.

The procedure of the two-stage column generation method is illustrated in Algorithm \ref{algo_tscg}.

\begin{algorithm}[!t]
\label{algo_tscg}
 	\caption{Two-Stage Column Generation (TSCG)}
     \SetKwInOut{Input}{Input}
     \SetKwInOut{Output}{Output}

     \SetKw{Break}{break}

     \Input{network $G$, crew $R$}
     \Output{minimum cost $c^*$, optimal roster $\{r: \gamma_r^* | r \in R\}$}

     \tcc{Network flow optimization}
     Construct initial path set $\tilde{\Gamma}$ with $G$ using Algorithm S.2 \;
     
     \While{True}{
          Obtain dual varaibles $\mu_S, \nu_k$ by solving $\mathbf{RLMP}$ with $\tilde{\Gamma}$ using solver\;
          Obtain shortest path $\gamma$ and reduced cost $\sigma$ by solving $\mathbf{PP}$ with $G, \mu_S, \nu_k$ using Algorithm S.3\;
          
          \If{$\sigma \ge 0$}{
               \Break \;
          }
          $\tilde{\Gamma} \gets \tilde{\Gamma} \cup \{\gamma\}$ \;
     }
     Obtain objective value $c_1^*$ and path set $\Gamma^*$ by solving $\mathbf{RIP}$ with $\tilde{\Gamma}$ using solver;

     \tcc{Assignment optimization}

     \For{$\gamma \in \Gamma^*$}{
          \For{$r \in R$}{
               Calculate $c^r_\gamma$ according to Eq. (\ref{cost_matrix}) \;
          }
     }
     Obtain objective value $c_2^*$ and optimal roster $\{r: \gamma_r^* | r \in R\}$ by solving $\mathbf{P_2}$ with $\Gamma^*$, $c_\gamma^r$ using solver\;
     $c^* \gets c_1^* + c_2^*$ \;
     
\end{algorithm}

\subsubsection{Network Flow Optimization}
\label{section_P1}
Problem $\mathbf{P_1}$ is first introduced, which can be viewed as a variant of the minimum-cost network flow problem aiming to search for the optimal group of valid paths.
Focusing on the path-based decision variables $x_\gamma$ and the associated constraints,
problem $\mathbf{P_1}$ is written as follows:
\begin{align}
 \mathbf{P_1}:\mathrm{Minimize} \ &\sum_{\gamma \in \Gamma}\sum_{u \to v \in A_\gamma}c(u, v) x_\gamma  + \sum_{k\in K}\lambda_k \label{obj_first_stage} \\
\mathrm{subject \ to} \ & \mathrm{constraints} \ (\ref{c1}), (\ref{c2}), (\ref{c5})-(\ref{c8}) \nonumber
\end{align}

Column generation is adopted here to handle the excessive number of variables.
As illustrated in Fig. \ref{pipeline}, after constructing an initial feasible solution, 
column generation begins an iterative procedure between the relaxed linear master problem ($\mathbf{RLMP}$) and the pricing problem ($\mathbf{PP}$).
This procedure continues until the $\mathbf{PP}$ fails to identify a variable that can be added to improve the solution quality.
Finally, the $\mathbf{RLMP}$ is converted into the restricted integer problem ($\mathbf{RIP}$) with the current set of variables to obtain the optimal integer solution.

The tailored pulse algorithm for generating constraint shortest paths, the initial solution construction, $\mathbf{RLMP}$, $\mathbf{PP}$ and $\mathbf{RIP}$ are next introduced as follows:

Since multiple constraints are involved in problem $\mathbf{P_1}$, the task of finding paths with minimum cost naturally becomes a constrained shortest path problem\cite{beasley1989algorithm}.
The pulse algorithm\cite{lozano2013exact,lozano2016exact} is one widely used labeling method for finding constrained shortest paths
based on the idea of propagating labels recursively through a network.
This work utilizes the pulse algorithm to generate valid paths, and design tailored pruning strategies 
and traversal method to improve searching efficiency.
Details of the pulse algorithm are illustrated in Section S.III-A in the supplementary material.

Before solving $\mathbf{RLMP}$,  an initial path set is first generated, 
denoted as $\tilde{\Gamma} \subseteq \Gamma$, which constitutes one feasible solution to problem $\mathbf{P_1}$.
To this end, the pulse algorithm is leveraged to search for the constrained shortest path for each crew member, building an initial path set $\tilde{\Gamma}$ with size $n_r$.
During the shortest path search for different crew members, certain types of arc costs of the time-space network should be adjusted to meet the operational rules.
Details of the initial path set construction algorithm are illustrated in Section S.III-B.

\label{rlmp}
With the generated initial subset $\tilde{\Gamma}$, 
the problem $\mathbf{RLMP}$ is written as follows:
\begin{align}
\mathbf{RLMP}: \mathrm{Minimize} \ &\sum_{\gamma \in \tilde{{\Gamma}}}\sum_{u \to v \in A_\gamma}c(u, v) x_\gamma +\sum_{k\in K}\lambda_k & \label{r_object} \\
\mathrm{subject \ to} \ &\sum_{\gamma \in \tilde{\Gamma}}\sum_{q \in S}m_{\gamma}^{q}x_\gamma \le \sum_{q \in S}\left |P_q \right |, \ \ \forall S \in \mathbf{S}&\label{r_c1}\\
& \sum_{\gamma \in \tilde{\Gamma}}\kappa_{\gamma}^k x_\gamma \le 1, \ \ \forall k \in K &\label{r_c2} \\
& x_\gamma \ge 0, \ \ \forall \gamma \in \tilde{\Gamma} &\label{r_c3}
\end{align}
Note that constraints (\ref{r_c3}) are modified to non-negative bounds for linear relaxation.
The reduced cost for each decision variable is next calculated. Let $\mu_S$  and $\nu_k$ as the dual variables associated with constraints (\ref{r_c1}) and (\ref{r_c2}), respectively. 
The reduced cost $\sigma_\gamma$ for decision variable $x_\gamma$ is then calculated as follows:
\begin{equation}
\label{rc}
\sigma_\gamma = \sum_{u \to v \in A_\gamma}c(u, v) - \sum_{S \in \mathbf{S}}\sum_{q \in S}m_{\gamma}^{q} \mu_{S} - \sum_{k \in K}\kappa_\gamma^k\nu_k
\end{equation}
Mathematically, if the reduced costs $\sigma_\gamma$ for all $\gamma \in \Gamma \setminus \tilde{\Gamma}$ are non-negative,
the optimality of the linearized $\mathbf{P_1}$ is met.
Otherwise, the vairiable $x_\gamma$ 
with the minimum negative reduced cost can be selected for inclusion in $\tilde{\Gamma}$
, as it offers the greatest potential to improve the objective function.

The pricing problem aims at searching for the valid path with minimum reduced cost.
To incorporate the dual variable terms,
let $\nu_k$ denote the costs of the corresponding train arcs in $A$, 
and include $\mu^q_\gamma$ as extra cost on qualification selection.
Then, the pricing problem is described based on the constraint shortest path formulation. 
Let $z_{u,v} \in \{0, 1\}$ denote the decision variable indicating whether the arc $u \to v$ is selected in the shortest path.
The problem $\mathbf{PP}$ is then formulated as follows:
\begin{align}
\mathbf{PP}: \underset{q \in Q}{\mathrm{Minimize}} \ &\sum_{u \to v \in A}c(u, v) z_{u,v} - \sum_{S \in \mathbf{S}}\delta(q \in S)\mu_{S} & \label{pp_object} \\
\mathrm{subject \ to} \ &\sum_{\{u | v_{sc} \to u \in A\}}z_{v_{sc}, u} = 1 & \label{pp_c1}\\
& \sum_{\{u | u \to v_{sk} \in A\}}z_{u, v_{sk}} = 1 & \label{pp_c2}\\
& \sum_{\{u | u \to v \in A\}}z_{u, v} = \sum_{\{u' | v \to u' \in A\}}z_{v, u'}, \nonumber\\
& \quad \quad \quad \quad \quad \quad \quad \forall v \in V \setminus\{v_{sc}, v_{sk}\} & \label{pp_c3} \\
& n^{si} \le n_d - n_{df} \label{pp_c4}  \\
& n^{dh} \le n_{tf}  \label{pp_c5} \\
& n^{ml, d} = 1, \  \ \forall d \in D & \label{pp_c6} \\
& z_{u, v} \in \{0, 1\}, \ \ \forall u \to v \in A & \label{pp_c7}
\end{align}
The objective function (\ref{pp_object}) is minimized over all qualifications, based on which the arc costs are revised.
Constraints (\ref{pp_c1})-(\ref{pp_c3}) are the typical flow conservation constraints.
Problem $\mathbf{PP}$ can be solved utilizing the pulse algorithm, as described in Algorithm S.1 in the supplementary material.
The resulting objective value for $\mathbf{PP}$ represents the minimum reduced cost based on the current $\tilde{\Gamma}$.
If the value is non-negative, then the termination condition for column generation is met. 
Otherwise, the resulting shortest path is added to $\tilde{\Gamma}$ for $\mathbf{RLMP}$.

After the termination condition is met, 
the $\mathbf{RLMP}$ problem is converted to the integer programming version $\mathbf{RIP}$ by modifying constraints (\ref{r_c3}) to boolean limits.
The problem can be then solved utilizing integer programming solvers.
Based on the integer solution, the optimal group of paths are selected, denoted as $\Gamma^*$.

\subsubsection{Assignment Optimization}
\label{section_P2}
The second stage optimization focuses on minimizing the preference penalties
with the selected path set $\Gamma^*$.
Since the existance of a perfect matching between the selected paths and crew is guaranteed 
by satisfying constraints (\ref{c1}) in problem $\mathbf{P_1}$,
problem $\mathbf{P_2}$ can be formalized as a standard assignment problem.
Let $c^r_{\gamma}$ represent the preference penalty for crew member $r$ to implement path $\gamma$:
\begin{equation}
\label{cost_matrix}
c^r_{\gamma}=
\begin{cases}
     +\infty,  \quad \quad \quad \quad \quad \quad \text{if} \ \sum_{q \in Q} m_{\gamma}^q \delta(q \subseteq q_{r}) < 1\\
     \sum_{o \in O_{\gamma}}\delta(o \notin O^{r})\lambda_o,  \quad \quad \quad \text{otherwise}
\end{cases}
\end{equation}
Problem $\mathbf{P_2}$ is then presented as follows: 
\begin{align}
\mathbf{P_2}: \mathrm{Minimize} \ &\sum_{r \in R}\sum_{\gamma \in \Gamma^*}c^r_{\gamma}y_{\gamma}^{r} & \label{p2_object} \\
\mathrm{subject \ to} \ &\sum_{r \in R}y_{\gamma}^{r} = 1, \ \ \forall \gamma \in \Gamma^* & \label{p2_c1}\\
&\sum_{\gamma \in \Gamma^*}y_{\gamma}^{r} = 1, \ \ \forall r \in R & \label{p2_c2}\\
& y_{\gamma}^{r} \in \{0, 1\}, \ \ \forall r \in R,\forall \gamma \in \Gamma^* &  \label{p2_c3}
\end{align}
Constraints (\ref{p2_c1})-(\ref{p2_c2}) are the agent and task constraints for a typical assignment problem.

Combining the resulting objective value of $\mathbf{P_1}$ and $\mathbf{P_2}$, the optimized total cost for problem $\mathbf{P}$ is obtained.
The final roster is generated utilizing the optimal path set $\Gamma^*$ and the assignment results.

\subsection{Fast Path Adjustment Heuristic}
\label{section_algorithm_fast}

In the context of real-time replanning problems, where rapid decision-making is a critical factor \cite{cacchiani2014overview}, this work opts for heuristic algorithms over solvers and exact methods 
as they can efficiently provide near-optimal solutions within a shorter solving time.
For solving problem $\mathbf{RP}$, this work designs a network path adjustment heuristic on the basis of the hierarchical network and restricted subnetwork construction.

Inspired by \cite{caprara1997algorithms,zhou2021integrated}, 
this work adopts a greedy strategy to address the challenge of varying duty frames and work lines among crew members during replanning.
The core idea is to prioritize assigning tasks to the crew member with the highest capacity to handle tasks available at the replanning moment, enabling the cost-minimizing path search to be conducted first.
Specifically, the crew memberswho are actively working at the replanning start time $\bar{t}$ are first sorted in descending order of their remaining working hours for the replanning day $\bar{d}$.
Next, crew members who are not on duty at time $\bar{t}$ but are scheduled to sign in later in day $\bar{d}$ are sorted by their sign-in time in ascending order and place them after the crew members sorted in the first step.
Finally, following the sorted order, the constrained shortest path $\omega^*$ is obtained using the pulsing algorithm on the restricted subnetwork $G^{\gamma^*_r}$ constructed for each crew member $r$ to determine their new roster.

The procedure of the fast path adjustment heuristic is illustrated in Algorithm \ref{algo_fph}.

\begin{algorithm}
\label{algo_fph}
\caption{Fast Path Adjustment Heuristic (FPAH)}
\SetKwInOut{Input}{Input}\SetKwInOut{Output}{Output}

\Input{network $G$, original roster $\{r: \gamma_r^* | r \in R\}$, replanning settings $\bar{d}$, $\bar{t}$, $\bar{K}$}
\Output{replanned roster $T=\{r: \omega_r^* | r \in R\}$, minimum cost $c^*$}

     Revise the train arcs in $G$ with new task set $\bar{K}$ \;
     Determine the replanned path set $\bar{\Gamma} \subseteq \{\gamma_r^* | r \in R\}$ and $u_\gamma, t_{v_\gamma}$ for each $\gamma \in \bar{\Gamma}$ \;
     Initialize empty arrays $A, A_1, A_2$ and empty roster $T$\;
     $c^* \gets 0$\;
     \For{$\gamma_r^* \in \bar{\Gamma}$}{
          \lIf{$a_{u_{\gamma_r^*}} < \bar{t}$}{
               Append $r$ to $A_1$
          }
          \lElse{
               Append $r$ to $A_2$
          }
     }
     Sort $A_1$ descendingly by $a_{u_{\gamma_r^*}}+H-\max(\bar{t}, t_{v_\gamma})$ \;
     Sort $A_2$ ascendingly by $a_{u_{\gamma_r^*}}$ \;
     $A \gets \text{concat}(A_1, A_2)$ \; 
     \For{$r \in A$}{
          Construct restricted subnetwork $G^{\gamma_r^*}$ \;
          Obtain shortest path $\omega_r^*$ and objective value $c^*_r$ with $G^{\gamma_r^*}$ using Algorithm S.1 \;
          Revise train arc costs in $G$ based on $\omega_r^*$ using Eq.(S.5) \; 
          Append $\{r:\omega_r^*\}$ to $T$\;
          $c^* \gets c^* + c^*_r$ \;
     }
\end{algorithm}

\section{Numerical Experiments}
\label{section_study}

\subsection{Data and Benchmarks}

\begin{figure}
     \centering
     \includegraphics[height=0.33\textheight]{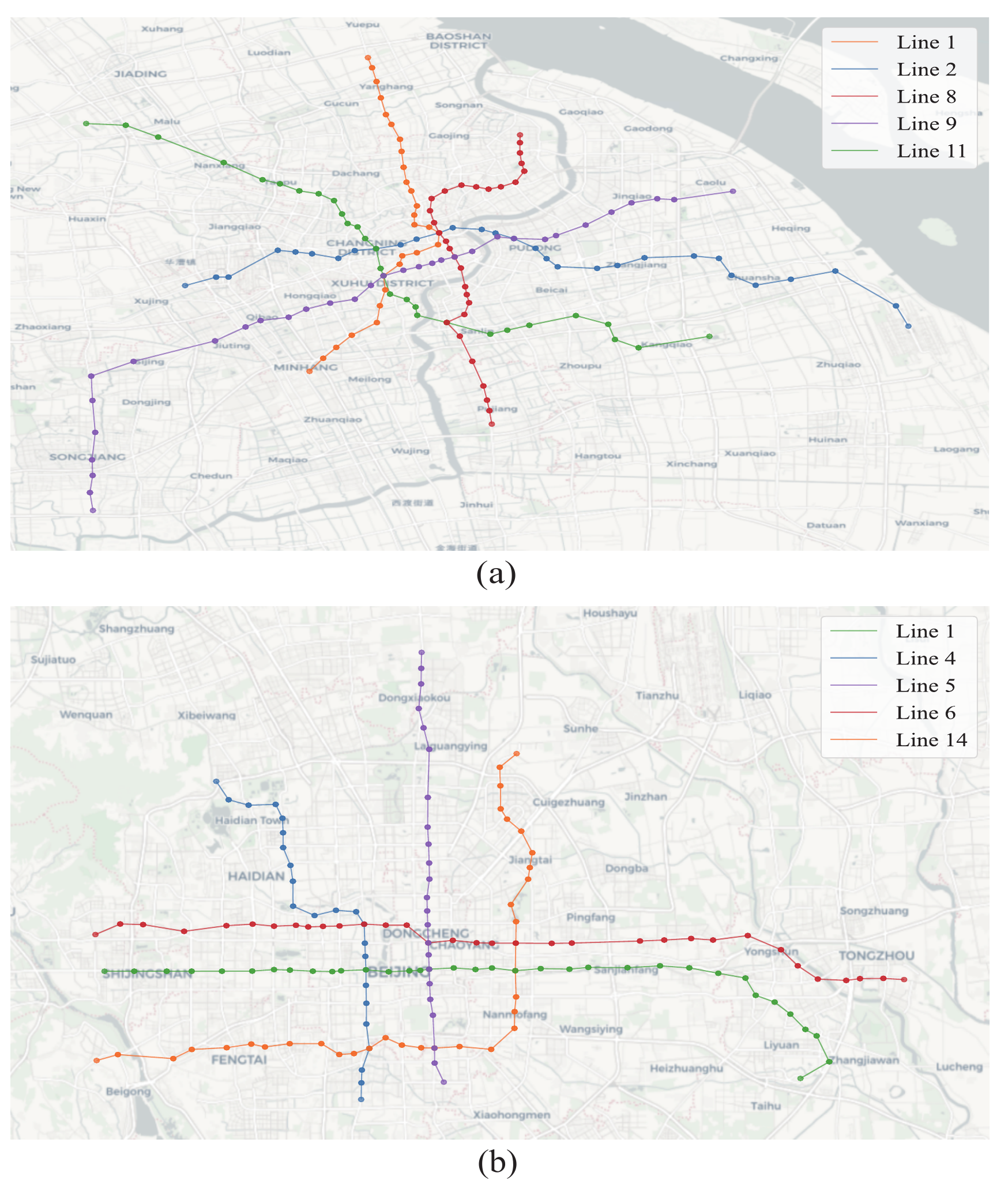}
     \caption{
          The network topology of the studied multi-line metro systems: (a) Shanghai Metro and (b) Beijing Metro.
     }
     \label{metro_topo}
\end{figure}

The model and algorithms are validated using real-life timetable data from the two largest metro systems globally in terms of system length and annual ridership \cite{wiki_metro}: the Shanghai Metro and the Beijing Metro.
The timetable data is sourced from the official metro websites of the three aforementioned cities \cite{shanghai_metro}, and follows the information published as of December 12, 2024.
For computational convenience, five representative high-traffic main lines are selected from each metro system as the focus of this study.
Fig. \ref{metro_topo} shows the topology of the two metro networks considered in the numerical experiments.
Consistent with the normal metro operating hours, this work considers train service tasks scheduled between 5:00 and 24:00 for each day.
The duration of train service tasks in each line is estimated as the average of the first and last train operating times in the timetable.
To simulate potential discrepancies between actual train departure times and the scheduled timetable during real-time operations, random temporal perturbations are introduced when generating train service tasks. Specifically, for each task $k$, a random integer between -1 and 1 is added to both the departure time $t_k$ and the arrival time $t_k'$.

In terms of algorithm comparison, due to the path-specific constraints in problem $\mathbf{P}$ and non-convex penalty terms in the objective, it is challenging to apply Lagrange relaxation or other exact methods for problem solving. 
Furthermore, the stringent temporal precision required in metro planning renders railway and airline crew planning algorithms unsuitable for direct adaptation and comparison.
As a result, two heuristic algorithms commonly used in single-line metro crew planning are selected and extended to multi-line versions to serve as benchmarks.
Specifically, the following two heuristic algorithms are considered:

\textbf{{Leg Generation Heuristic}}:
A leg refers to a list of consecutive tasks or actions performed by the same driver within one duty.
Similar to \cite{zhou2021integrated}, the presented leg generation heuristic (LGH) follows a time-sequential strategy to construct legs for each crew member, prioritizing the inclusion of as many tasks as possible while ensuring compliance with the working time rule.
Specifically, for each crew member requiring leg assignment, the working days are first ranked based on the remaining number of tasks, and select the day with the highest task count for leg construction. Within the qualified task set for this crew member, a leg comprising the maximum number of consecutive tasks that also satisfies the operational rules is assigned as the duty for this working day.
LGH can identify solutions within a very short time without the need for additional model construction.
The procedure for LGH is illustrated in Algorithm S.4 in the supplementary material.

\textbf{{Sequential Path Heuristic}}:
The sequential path heuristic (SPH) searches for the constrained shortest path for each crew member sequentially, a strategy similarly applied in \cite{caprara1997algorithms,zhou2022metro}. 
For each crew member, the algorithm adopts a greedy strategy for working line selection given the qualified line set. 
To handle path compatibility among different crew members, 
the heuristic follows the arc cost revision method expressed in Eq.(S.5). Note that this heuristic is implemented on the basis of the constructed hierarchical time-space network model.
The procedure for SPH is illustrated in Algorithm S.5 in the supplementary material.

All numerical experiments are implemented in Python,
and the subproblems $\mathbf{RLMP}, \mathbf{RIP}$ and $\mathbf{P_2}$ are solved using the Gurobi solver\cite{gurobi2021gurobi}.
The computer processor is an Intel(R) Xeon(R) Gold 5118 CPU, 2.30 GHZ, with 64.0 GB RAM.

\subsection{Crew Planning Performance Analysis}

The performance of crew planning is tested on a three-day planning horizon, utilizing timetable data from Monday to Wednsday. 
Taking understaffing into account, this work sets up multiple crew sizes $n_r$ in different test cases.
For heterogeneous crew generation, qualifications are assigned randomly following a uniform distribution. 
Specifically, this work assigns 40\% of the crew to be qualified for two randomly selected lines, while the remaining 60\% members are restricted to one random metro line.
The preferred depots $O_r$ for each crew member are set to two randomly selected depots from those located along the qualified lines. The remaining parameter settings are summarized in Table \ref{params}.
The optimization objective $Obj$, representing the total operational cost, is set as the first performance metric.
Considering the occurrence of understaffing cases, the task coverage ratio $Cvg$ is selected, defined as the percentage of successfully performed train service tasks from the timetable, as the second performance metric.
To ensure statistical validity, each test instance is simulated 20 times, and the average results are recorded.

\begin{table}
\centering
\footnotesize
\caption{Parameter Settings for Numerical Expermients}
\label{params}
\begin{tabular}{cc|cc}
\hline
Parameter & Value & Parameter & Value \\
\hline
$H$, $h$ & 540, 120 & $t_{min}, t_{max}$ & 530, 540 \\
$t_{rt}$, $t_{ml}$ &10, 45 & $t_{mb}, t_{me}$& 120, 420\\
 $t_{tf}$, $n_{df}$, $n_{tf}$ & 5, 1, 10 &  $t_{si}$, $t_{so}$ & 20, 20  \\
  $\lambda_k$, $\lambda_o$ &4($t_k - t_k'$), 50 &  $c_w, c_r$  & 1, 0.2 \\
\hline
\end{tabular}
\end{table}

\begin{figure}[!t]
     \centering
     \includegraphics[height=0.26\textheight]{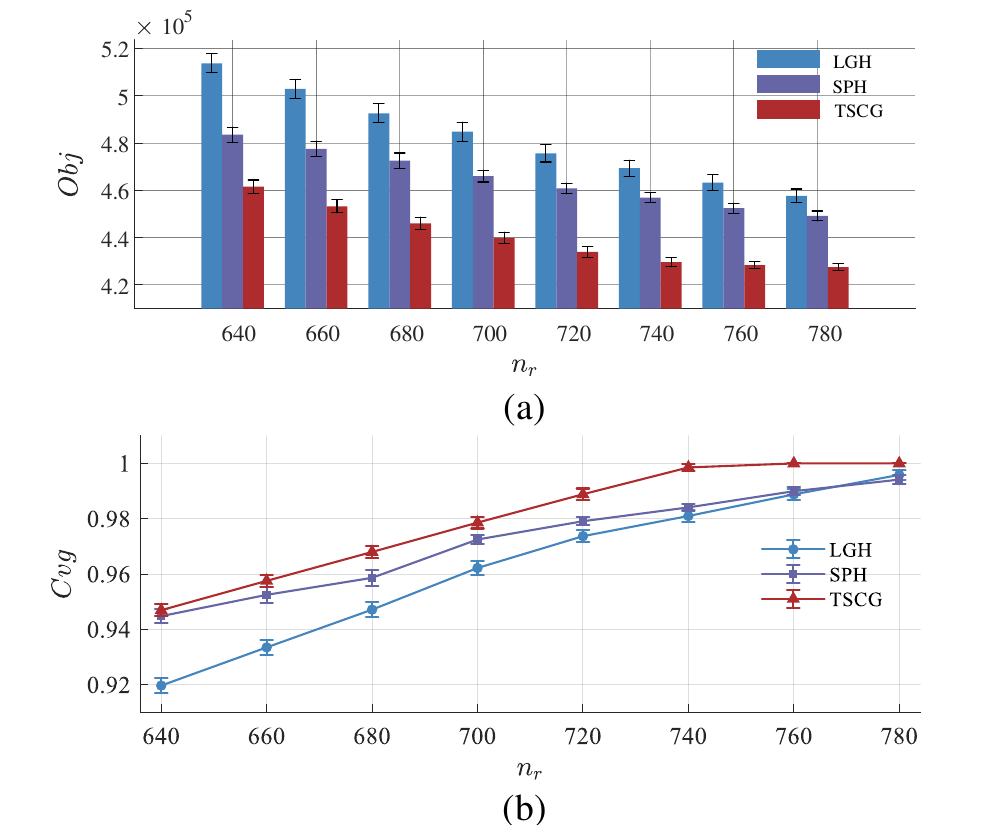}
     \caption{
          Crew planning performance comparison of TSCG and heuristic benchmarks in terms of (a) objective value and (b) task coverage ratio using real data from the Shanghai Metro, with $n_l = 3$. Error bars represent the standard deviation of the data.
     }
     \label{perf}
\end{figure}

\begin{figure}[!t]
     \centering
     \includegraphics[height=0.17\textheight]{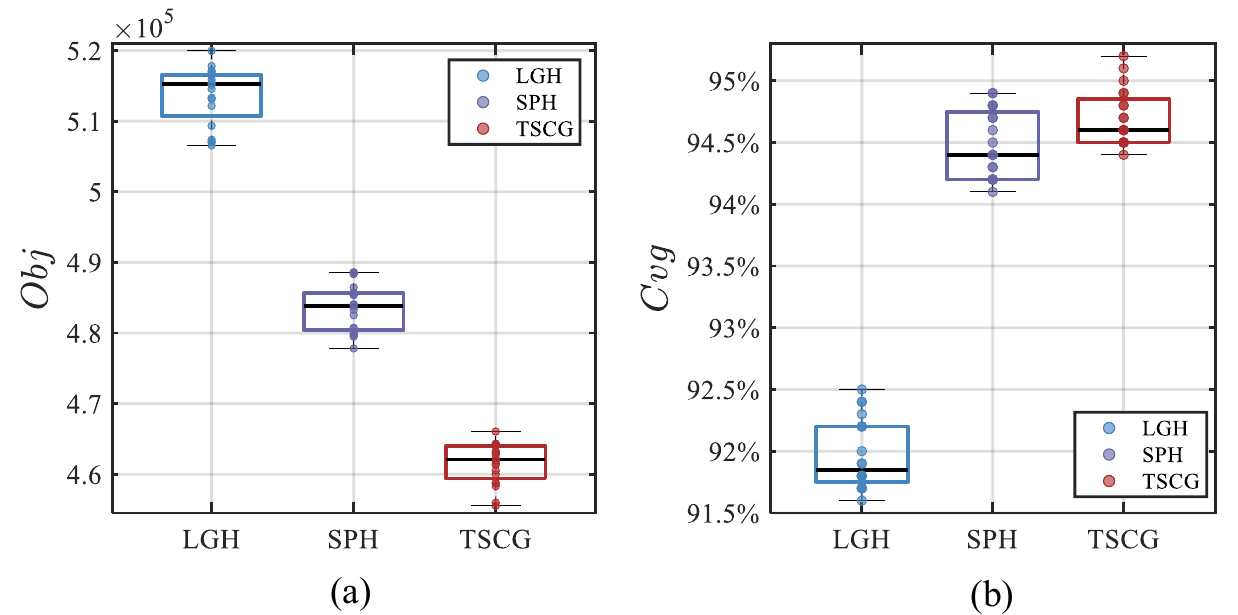}
     \caption{
    Box-plot comparison of TSCG and heuristic benchmarks' crew planning performance in terms of (a) objective value and (b) task coverage ratio using data from the Shanghai Metro, with $n_l=3$ and $n_r=640$.     
     }
     \label{perfboxplot}
\end{figure}

Tables \ref{perf_sh} and \ref{perf_bj} display the experimental results for Shanghai Metro and Beijing Metro. 
It can be observed that the proposed TSCG method outperforms the two heuristic algorithms across both evaluated metrics. 
Among the Shanghai test cases,
TSCG achieves up to an 11.21\% reduction in total operational cost and a 3.14\% improvement in task coverage ratio compared to the LGH.
When compared with SPH, which also leverages HTSN model, TSCG offers a modest improvement of less than 2\% in task coverage but achieves a 5\% to 6\% reduction in total cost, as depicted in Figure \ref{perf}. 
This improvement is attributed to the advantages of column generation in solving multi-commodity network flow problems, enabling the identification of more efficient and compact crew action sequences. 
Comparing the two heuristic algorithms, while LGH generates feasible solutions in a remarkably short time, the solution quality significantly lags behind SPH.
This highlights the considerable advantage of using HTSN and shortest-path-based optimization in crew planning tasks.
Figure \ref{perfboxplot} illustrates the performance variations under train departure time disturbances. 
It can be observed that the TSCG exhibits minimal fluctuations in both metrics, demonstrating resilience to operational pertubations.

\begin{table*}
     \caption{Crew Planning Performance Analysis Using Shanghai Metro Data}
     \centering
     \label{perf_sh}
     \begin{tabular}{cccccccccccc} 
\toprule
\multicolumn{1}{l}{} & \multicolumn{1}{l}{} & \multicolumn{1}{l}{} & \multicolumn{3}{c}{LGH}          & \multicolumn{3}{c}{SPH}      & \multicolumn{3}{c}{TSCG}  \\ 
\cmidrule(lr){4-6}\cmidrule(lr){7-9}\cmidrule(lr){10-12} 
$n_l$ & $n_k$    & $n_r$    & $Obj$        & $Cvg$   & Time(s) & $Obj$       & $Cvg$   & Time(s)  & $Obj$        & $Cvg$   & Time(s)    \\ 
\midrule
3  & 3,800 & 680   & 492,577.9 & 94.71\% & 0.11    & 472,523.1 & 95.86\% & 1,774.34 & 446,093.1 & 96.80\% & 2,370.36   \\
3  & 3,800 & 760   & 463,292.5 & 98.88\% & 0.14    & 452,445.8 & 99\%    & 2,031.24 & 428,326.5 & 100\%   & 2,858.20   \\
4  & 5,126 & 880   & 622,780.4 & 95.21\% & 0.16    & 594,447.6 & 96.62\% & 2,499.49 & 557,109.4 & 97.59\% & 4,510.43   \\
4  & 5,126 & 960   & 595,866.3 & 98.44\% & 0.16    & 573,509.5 & 99.14\% & 2,663.46 & 543,404.0    & 100\%   & 5,154.21   \\
5  & 6,174 & 1,060 & 751,636.7 & 94.90\% & 0.14    & 709,208.4 & 97.06\% & 2,660.17 & 667,370.8 & 97.88\% & 8,485.04   \\
5  & 6,174 & 1,140 & 725,568.2 & 97.56\% & 0.14    & 693,029.8 & 98.70\% & 2,789.05 & 652,705.7 & 100\%   & 8,969.84  \\
\bottomrule
\end{tabular}
\end{table*}

\begin{table*}
     \caption{Crew Planning Performance Analysis Using Beijing Metro Data}
     \centering
     \label{perf_bj}
     \begin{tabular}{cccccccccccc} 
\toprule
\multicolumn{1}{l}{} & \multicolumn{1}{l}{} & \multicolumn{1}{l}{} & \multicolumn{3}{c}{LGH}          & \multicolumn{3}{c}{SPH}      & \multicolumn{3}{c}{TSCG}  \\ 
\cmidrule(lr){4-6}\cmidrule(lr){7-9}\cmidrule(lr){10-12} 
$n_l$ & $n_k$    & $n_r$    & $Obj$        & $Cvg$   & Time(s) & $Obj$       & $Cvg$   & Time(s)  & $Obj$        & $Cvg$   & Time(s)    \\ 
\midrule

3                      & 4,638                   & 700                    & 526,862.3                    & 90.68\%                       & 0.12                      & 462,675.4                           & 96.37\%                              & 2,010.32                          & 438,690.3                    & 96.84\%                       & 3,521.31                   \\
3                      & 4,638                   & 780                    & 498,151.3                    & 95.18\%                       & 0.12                      & 450,678.4                           & 98.67\%                              & 2,172.58                          & 426,545.2                    & 100\%                         & 4,013.17                    \\
4                      & 6,062                   & 960                    & 690,352                      & 92.77\%                       & 0.15                      & 621,451.8                           & 97.17\%                              & 2,814.02                          & 589,447.7                    & 97.66\%                       & 5,323.5                    \\
4                      & 6,062                   & 1,040                   & 664,515.2                    & 96.01\%                       & 0.18                      & 611,383.2                           & 98.63\%                              & 2,970.2                           & 578,056.5                    & 100\%                         & 4,841.22                   \\
5                      & 7,180                   & 1,180                   & 813,596.6                    & 94.22\%                       & 0.15                      & 742,988.9                           & 97.64\%                              & 3,105.91                          & 709,998.7                    & 98.22\%                       & 10,959.74                   \\
5                      & 7,181                   & 1,260                   & 791,196.3                    & 96.61\%                       & 0.16                      & 734,107.9                           & 98.77\%                              & 3,272.07                          & 699,657                      & 100\%                         & 17,369.46                  \\

\bottomrule
\end{tabular}
\end{table*}

\subsection{Crew Replanning Performance Analysis}

To evaluate the performance of the proposed crew replanning method, this work first designs a disruption scenario to simulate the changes in task demands during replanning.
This work focuses on passenger surges, a common type of metro disruption. 
Specifically, this work considers passneger surges in Shanghai Metro that occur on Line 1 during peak hours.
The replanning day is set as the second day of a three-day planning horizon.
Both morning and evening peak traffic surges are simulated, and the replanning processes are initiated at 6:30 and 16:30, i.e., $\bar{t}=90$ and $\bar{t}=690$, respectively. 
To reflect timetable adjustments that accommodate the increased demand, 
the train departure intervals are shortened to 2 minutes in the 1 to 3 hours after the start of crew replanning on Line 1.  
The canceling penalty multiplier is also tripled for these replanned urgent tasks to prioritize the satisfaction of urgent demands induced by the disruption.

Due to the requirements of spatiotemporal continuity in metro crew replanning, existing railway replanning models cannot be directly applied.
For comparison, the leg generation heuristic is extended for the replanning scenario, termed LGH-R. Specifically, LGH-R incorporates a greedy prioritization of urgent tasks during the leg generation process, as detailed in Algorithm S.6 in the supplementary material. 
Additionally, ablation studies are conducted by introducing no-deadhead constraints into the FPAH method, termed FPAH-N, 
to analyze the impact of cross-line operations on replanning performance. 
To provide a more comprehensive assessment, an additional metric, $Cvg_u$, is introduced to measure the completion rate of urgent train service tasks during the replanning phase.

Tables \ref{reschedule_morning} and \ref{reschedule_night} illustrate the replanning results in response to morning and evening passenger surges, respectively.
The proposed FPAH method demonstates effective performance in terms of both operational cost control and task completion.
In the cases of morning passenger surges, FPAH reduces total costs by more than 25\% compared to LGH-R, while also improving the  overall task coverage rate by approximately 15\%. Notably, FPAH excels in the completion of urgent tasks, showing an increase of over 30\% compared to LGH-R. 
For evening surge scenarios, FPAH decreases total costs by 6\% to 10\% compared to LGH-R. Although the overall coverage rate improvement is less than 2\% over LGH-R, FPAH still enhances the completion rate of emergency tasks by 10\% to 15\%. 
The primary advantage of FPAH over LGH-R lies in its ability to conduct a global path search on the restricted subnetwork, thereby optimizing crew task allocation between emergency routes and other routes, particularly in scenarios with extended replanning periods. 
In the cases with morning surges, the inefficiencies observed with LGH-R indicate that the greedy approach to selecting urgent tasks could markedly compromise subsequent scheduling, resulting in diminished overall efficiency. In contrast, in the evening surge scenarios, where replanning initiates later, 
the reduced subsequent replanning search space means that employing a greedy approach could yield overall task completion rates comparable to those of FPAH.
These findings underscore the superiority of network flow path searching methodologies when addressing large-scale, heterogeneous task planning problems.

\begin{table*}
\centering
\caption{Crew Replanning Performance Analysis with Morning Traffic Surges in Shanghai Metro ($\bar{t}=90$)}
\label{reschedule_morning}
\begin{tabular}{ccccccccccccc} 
\toprule
\multicolumn{1}{l}{} & \multicolumn{1}{l}{} & \multicolumn{1}{l}{} & \multicolumn{3}{c}{LGH-R}          & \multicolumn{3}{c}{FPAH}      & \multicolumn{3}{c}{FPAH-N}  \\ 
\cmidrule(lr){4-6}\cmidrule(lr){7-9}\cmidrule(lr){10-12} 
$n_l$  & $\bar{n_k}$    & $\bar{n_r}$    & $Obj$        & $Cvg$ / $Cvg_u$ & Time(s) & $Obj$       & $Cvg$ / $Cvg_u$   & Time(s)  & $Obj$        & $Cvg$ / $Cvg_u$   & Time(s)    \\ 
\midrule
3    & 1,262 & 436  & 202,649.7 & 87.1\% / 68.9\% & 0.04  & 184,982.8 & 89.7\% / 80.9\% & 197.59 & 190,367.5 & 89.3\% / 70.7\% & 149.98  \\
3    & 1,262 & 468  & 194,572.6 & 89.1\% / 75.5\%  & 0.04  & 171,727.9 & 92.4\% / 93.1\% & 190.11 & 173,833.6 & 92.5\% / 84.6\% & 151.05  \\
4    & 1,685 & 559  & 242,993.2 & 88.5\% / 71.0\%   & 0.07  & 218,818.1 & 91.8\% / 87.1\%  & 247.45 & 223,317.6 & 91.5\% / 79.9\%  & 207.33  \\
4    & 1,685 & 602  & 238,638.7 & 90.0\% / 76.5\%  & 0.07  & 210,435.9 & 94.0\% / 92.8\%  & 260.22 & 213,187.8 & 93.7\% / 85.7\%  & 212.11  \\
5    & 2,022 & 673 & 284,264 & 88.7\% / 75.2\%  & 0.05  & 257,007.3 & 91.9\% / 92.8\%  & 271.8  & 264,631.6 & 91.9\% / 79\%  & 258.55  \\
5   & 2,022 & 714 & 281,068.1 & 90.2\% / 74.8\%   & 0.07  & 249,063.9 & 94.0\% / 94.3\%  & 293.5  & 255,204.1 & 93.8\% / 82.8\%  & 261.72  \\
\bottomrule
\end{tabular}
\end{table*}
\begin{table*}
\caption{Crew Replanning Performance Analysis with Evening Traffic Surges in Shanghai Metro ($\bar{t}=690$)}
\label{reschedule_night}
\centering
\begin{tabular}{ccccccccccccc} 
\toprule
\multicolumn{1}{l}{} & \multicolumn{1}{l}{} & \multicolumn{1}{l}{} & \multicolumn{3}{c}{LGH-R}          & \multicolumn{3}{c}{FPAH}      & \multicolumn{3}{c}{FPAH-N}  \\ 
\cmidrule(lr){4-6}\cmidrule(lr){7-9}\cmidrule(lr){10-12} 
$n_l$  & $\bar{n_k}$    & $\bar{n_r}$    & $Obj$        & $Cvg$ / $Cvg_u$ & Time(s) & $Obj$       & $Cvg$ / $Cvg_u$   & Time(s)  & $Obj$        & $Cvg$ / $Cvg_u$   & Time(s)    \\ 
\midrule
3  & 505  & 256  & 93,237    & 78.2\% / 89.8\%  & 0.02  & 80,431.3  & 83.6\% / 98.1\% & 35.48  & 96,829.8  & 82\% / 64.8\% & 33.7    \\
3   & 505  & 271  & 89,049.6    & 81.6\% / 90.1\%   & 0.02  & 74,675.6  & 87.6\% / 99.6\% & 39.12  & 88,657.5  & 87.4\% / 70.8\%  & 32.53   \\
4   & 657  & 318  & 116,146.1 & 78.5\% / 82.9\%  & 0.02  & 98,059.8  & 84.1\% / 96.2\%  & 47.81  & 114,387.8 & 83.3\% / 64.2\%  & 42.37   \\
4   & 657  & 346  & 109,580.6  & 83.5\% / 84\%  & 0.02  & 88,889.6  & 89.9\% / 99.2\%  & 51.09  & 104,382.4 & 89\% / 70.1\% & 40.84   \\
5   & 764  & 373 & 133,202.5 & 80.4\% / 77\%  & 0.02  & 113,945.3 & 85\% / 92.3\%     & 55.9   & 126,826.1 & 84.7\% / 66.8\%  & 47.58   \\
5   & 764  & 401 & 125,015.2 & 84.6\% / 81.4\%  & 0.03  & 104,617.3 & 89.7\% / 96.5\%  & 58.96  & 119,072.5 & 89\% / 69.9\%  & 52.08   \\

\bottomrule
\end{tabular}
\end{table*}

\begin{figure}
     \centering
     \includegraphics[height=0.37\textheight]{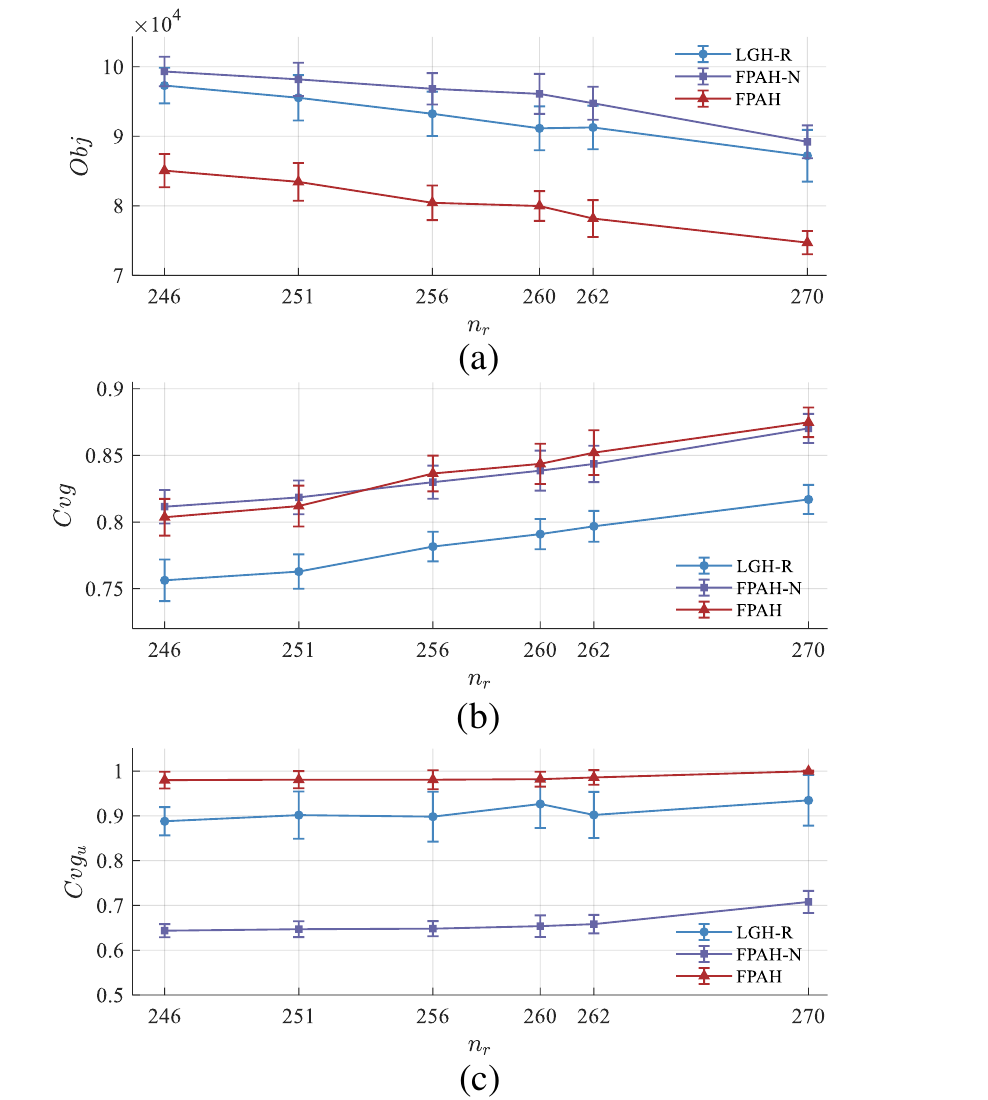}
     \caption{
          Crew replanning performance comparison of FPAH, FPAH-N and heuristic benchmark in terms of (a) objective value, (b) overall task coverage and (c) urgent task coverage during evening passenger surges in the Shanghai Metro, with $n_l = 3$. Error bars represent the standard deviation of the data.
     }
     \label{reschedulePDF}
\end{figure}

The ablation studies highlight the critical role of cross-line operations in multi-line crew replanning, particularly in improving the completion rates of urgent tasks. Compared to FPAH, FPAH-N exhibits a decrease in $Cvg_u$ by approximately 10\% in morning scenarios and over 20\% in evening scenarios.
While the overall task coverage ratio  remains comparable between FPAH and FPAH-N, the heightened canceling penalty for urgent tasks leads to higher operational costs for FPAH-N. 
In the evening passenger surge cases, as depicted in Figure \ref{reschedulePDF}, the operational cost of FPAH-N increases by over 10\% compared to FPAH and even exceeds that of LGH-R by approximately 1\%.
Such results highlight the efficiency gains enabled by incorporating cross-line operations into multi-line crew replanning, with particularly pronounced improvements in urgent task completion rates.

In terms of execution time, FPAH operates within one minute for evening replanning scenarios and keeps computation time under five minutes for morning replanning cases, thereby satisfying the potential demands for real-time computation.

\subsection{Optimality Analysis}

The TSCG algorithm employs a two-stage approach to solve problem $\mathbf{P}$ after model decomposition, which may introduce potential optimality losses. 
To assess this, this work compares the results obtained using TSCG with those from a direct one-step approach using the Gurobi solver 
after determining the candidate set $\tilde{\Gamma}$. Table \ref{optimality_analysis} presents a detailed comparison of the two methods, 
including solution quality and computational time after fixing $\tilde{\Gamma}$.
As shown in the table, for planning scenarios with three or four lines, the task coverage ratio obtained by TSCG is nearly identical to that of Gurobi, 
with the total cost differing by less than 0.2\%, indicating minimal performance loss. 
For test cases with five lines, the computational cost of one-step solving with Gurobi becomes excessively high compared with TSCG,
and the computational RAM required exceeds the capacity of the computer hardware.

These gaps arise because, compared with solving the decomposed subproblems $\mathbf{RIP}$ and $\mathbf{P_2}$,
the number of decision variables in the one-step approach increases from $O(|\tilde {\Gamma}|)$ to $O(|\tilde{\Gamma}|\cdot n_r)$,
and the number of constraints grows from $O(|\tilde{\Gamma}| + n_k)$ to $O(|\tilde{\Gamma}| \cdot n_q \cdot n_r + n_k)$.
Consequently, as the planning scale increases, the rapid growth of $n_r$ substantially raises the computational complexity of the solver.
These results demonstrate that the proposed TSCG algorithm can efficiently provide solutions with acceptable optimality losses, especially for large-scale planning problems.

\begin{table*}
\caption{Optimality Analysis Comparing TSCG and One-Step Solution Using Gurobi Solver}
\label{optimality_analysis}
\centering
\begin{tabular}{ccccccccccc} 
\toprule
\multicolumn{1}{l}{} & \multicolumn{1}{l}{} & \multicolumn{1}{l}{} & \multicolumn{3}{c}{TSCG}          & \multicolumn{3}{c}{Gurobi}      & \multicolumn{2}{c}{Comparison}  \\ 
\cmidrule(lr){4-6}\cmidrule(lr){7-9}\cmidrule(lr){10-11} 
$n_l$  & $n_k$    & $n_r$    & $Obj$        & $Cvg$  & Time(s) & $Obj$       & $Cvg$   & Time(s)  & $Gap_{obj}$        & $Gap_{cvg}$ \\ 
\midrule
3  & 3,800 & 640  & 461,541.6 & 94.7\%  & 68.86   & 461,057.5         & 94.7\%             & 197.19  & 0.1\%              & 0\%                 \\
3  & 3,800 & 680  & 446,093.1 & 96.8\%  & 88.36   & 445,216.5         & 96.8\%             & 290.82  & 0.2\%              & 0\%                 \\
4  & 5,126 & 880  & 557,109.4 & 97.59\% & 141.43  & 557,076.4         & 97.59\%            & 685.4   & 0.01\%             & 0\%                 \\
4  & 5,126 & 920  & 548,940.6 & 99.15\% & 151.28  & 548,875.9         & 99.15\%            & 768.15  & 0.01\%             & 0\%                 \\
5  & 6,174 & 1,100 & 658,993.8 & 99.16\% & 293.95  & \textbackslash{} & \textbackslash{} & $>$2,000    & \textbackslash{} & \textbackslash{}  \\
\bottomrule
\end{tabular}
\end{table*}

\section{Discussion}
\label{section_conclusion}

A unified crew planning
and replanning optimization framework is proposed to address the special challenges
in multi-line metro crew operations.
The framework introduces a novel time-space network model, HTSN, which employs a 
tailored hierarchical structure to 
provide a unified representation of crew planning and replanning action spaces.
To address the heterogeneous qualifications and preferences of crew members in multi-line scenarios, 
A set of computationally efﬁcient constraints are derived, 
upon which integer programming formulations are presented for both planning and replanning.
Two efficient methods, TSCG and FPAH, are further designed leveraging constrained shortest paths in HTSN for problem solving.

Utilizing real-world data from Shanghai Metro and Beijing Metro,
it is demonstrated that the proposed TSCG and FPAH outperform other heuristic benchmarks by comparative numerical experiments.
The empirical results highlight that effectively designed time-space network structure significantly contribute to
identifying more efficient and compact crew action sequences, 
particularly in large-scale planning and replanning scenarios.
Results also show that the introduction of cross-line operations 
significantly enhances overall performance, 
especially in completing urgent tasks.
By emphasizing global optimization and multi-line coordination, 
this work demonstrates their key contribution to improving operational efficiency 
and enhancing resilience to disruptions in metro systems. 
These strategies are fundamental for the functioning and emergency management of smart cities, 
enabling responsive and adaptive transportation systems in urban environments.

Looking ahead, the framework can be further refined in two key areas. 
First, the current model simplifies workforce heterogeneity, 
considering only qualifications per metro line and depot preferences. 
Diverse crew attributes can be incorporated for a more realistic representation. 
Second, cross-line strategies across different metro operational stages will be explored, 
such as express train scheduling \cite{yang2020service} and timetabling \cite{zhang2024integrated}. 
Incorporating these strategies have the potential to further optimize multi-line crew planning and 
contribute to the global optimization of intelligent metro system operations.

\section*{\MakeUppercase{A}\MakeLowercase{cknowledgements}}
The author greatly appreciates the constructive comments and help 
from Dingding Han,  Wei Yang and Huiyu Zhi 
at Fudan University.


\bibliographystyle{IEEEtran}
\bibliography{main}

\vspace{-100pt}

\begin{IEEEbiography}[{\includegraphics[width=1in,height=1.25in,clip,keepaspectratio]{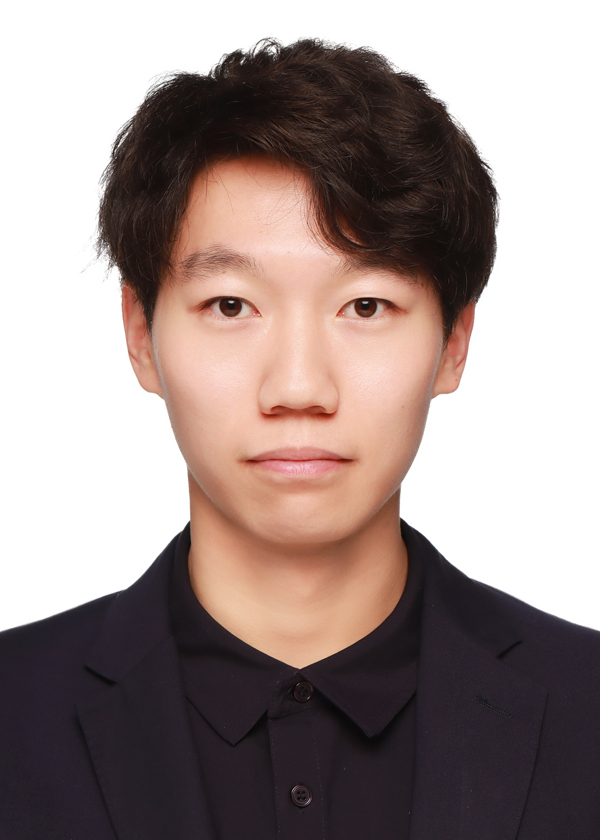}}]{Qihang Chen}
received the B.S. degree in intelligent science and technology from Fudan University, Shanghai, China, 
where he is currently pursuing the Ph.D. degree with the School of Information Science and Technology. 

His research interests include network science, discrete optimization, and their applications in intelligent transportation systems.
\end{IEEEbiography}

\end{document}